\documentclass[runningheads]{llncs}
\usepackage{cite}
\usepackage{graphicx,epstopdf,epsfig}
\graphicspath{{figures/}}
\DeclareGraphicsExtensions{.eps}
\usepackage{amsmath}
\usepackage{array}
\usepackage{fixltx2e}
\usepackage{url}
\usepackage{booktabs}
\usepackage{float}
\usepackage[caption=false,font=scriptsize,labelfont=sf,textfont=sf]{subfig}

\begin{document}
\title{
	A Visual Neural Network for Robust Collision Perception in Vehicle Driving Scenarios\thanks{Supported by the EU Horizon 2020 projects STEP2DYNA(691154) and ULTRACEPT(778062). Corresponding authors \email{\{qifu, syue\}@lincoln.ac.uk}}}
\titlerunning{Visual Collision Detection in Complex Dynamic Scenes}

\author{Qinbing Fu\inst{1,2}\orcidID{0000-0002-5726-6956} \and
Nicola Bellotto\inst{1}\orcidID{0000-0001-7950-9608} \and
Huatian Wang\inst{1}\orcidID{0000-0003-3589-9880} \and
F. Claire Rind\inst{3}\orcidID{0000-0002-8466-4597} \and
Hongxin Wang\inst{1}\orcidID{0000-0001-6184-4336} \and
Shigang Yue\inst{1,2}\orcidID{0000-0002-1899-6307}}
\authorrunning{Q. Fu et al.}
%
\institute{Lincoln Centre for Autonomous Systems (L-CAS), University of Lincoln, UK \and
School of Mechanical and Electrical Engineering, Guangzhou University, China \and
Institute of Neuroscience, Newcastle University, UK
}
\maketitle              
\begin{abstract}
This research addresses the challenging problem of visual collision detection in very complex and dynamic real physical scenes, specifically, the vehicle driving scenarios. 
This research takes inspiration from a large-field looming sensitive neuron, i.e., the lobula giant movement detector (LGMD) in the locust's visual pathways, which represents high spike frequency to rapid approaching objects. 
Building upon our previous models, in this paper we propose a novel inhibition mechanism that is capable of adapting to different levels of background complexity. 
This adaptive mechanism works effectively to mediate the local inhibition strength and tune the temporal latency of local excitation reaching the LGMD neuron. 
As a result, the proposed model is effective to extract colliding cues from complex dynamic visual scenes. 
We tested the proposed method using a range of stimuli including simulated movements in grating backgrounds and shifting of a natural panoramic scene, as well as vehicle crash video sequences. 
The experimental results demonstrate the proposed method is feasible for fast collision perception in real-world situations with potential applications in future autonomous vehicles.

\end{abstract}
\section{Introduction}
\label{Section: introduction}
Autonomous vehicles, though still in early stages of development, have demonstrated huge potential for shaping our future lifestyles and benefiting a variety of human activities. 
Before well serving the human society, there is one critical issue to solve -- trustworthy collision perception. 
Nowadays the number of fatalities by road crashes still remains high. 
To improve driving safety, the cutting-edge approaches for vehicle collision detection, such as radar, GPS-based methods and normal vision sensors, are often ineffective in terms of reliability, cost, energy consumption or size.

For ground vehicle collision detection, the most effective systems comprise automated collision avoidance with emergency steering and braking assistance, as well as active lane keeping systems, e.g. \cite{traffic-safety-2011}. 
The vast majority of vision-based methods implement object-and-scene segmentation, estimation or classification algorithms \cite{Vehicle-Review-2015}. 
The state-of-the-art visual sensors like RGB-D and event-driven cameras can provide vehicles with more abundant visual features compared to normal cameras. 
However, these solutions are either computationally costly or heavily reliant on the specific sensors.
A new type of reliable, low-cost, energy-efficient and miniaturised collision detection techniques is demanded for future autonomous vehicles.

Nature provides a rich source of inspiration for designing artificial visual systems for collision perception and avoidance (e.g. \cite{LGMD1-Glayer,LGMD2-BMVC,Fu-2018(LGMD1-NN),Nicolas-review,Serres-2017,Fu-2018(TAROS)}). 
Insects have compact visual brains that deal with motion perception. 
For instance, locusts can fly for a long distance in very dense swarms without collision; also nocturnal insects successfully forage in the forest at night free of collision. 
These naturally developed visual systems are perfect sensory models for collision detection and avoidance. 
In the future, vehicles with or without a driver should possess similar ability to navigate as effectively as animals do.
\begin{figure}[t]
	\vspace{-10pt}
	\centering
	\includegraphics[width=0.4\linewidth]{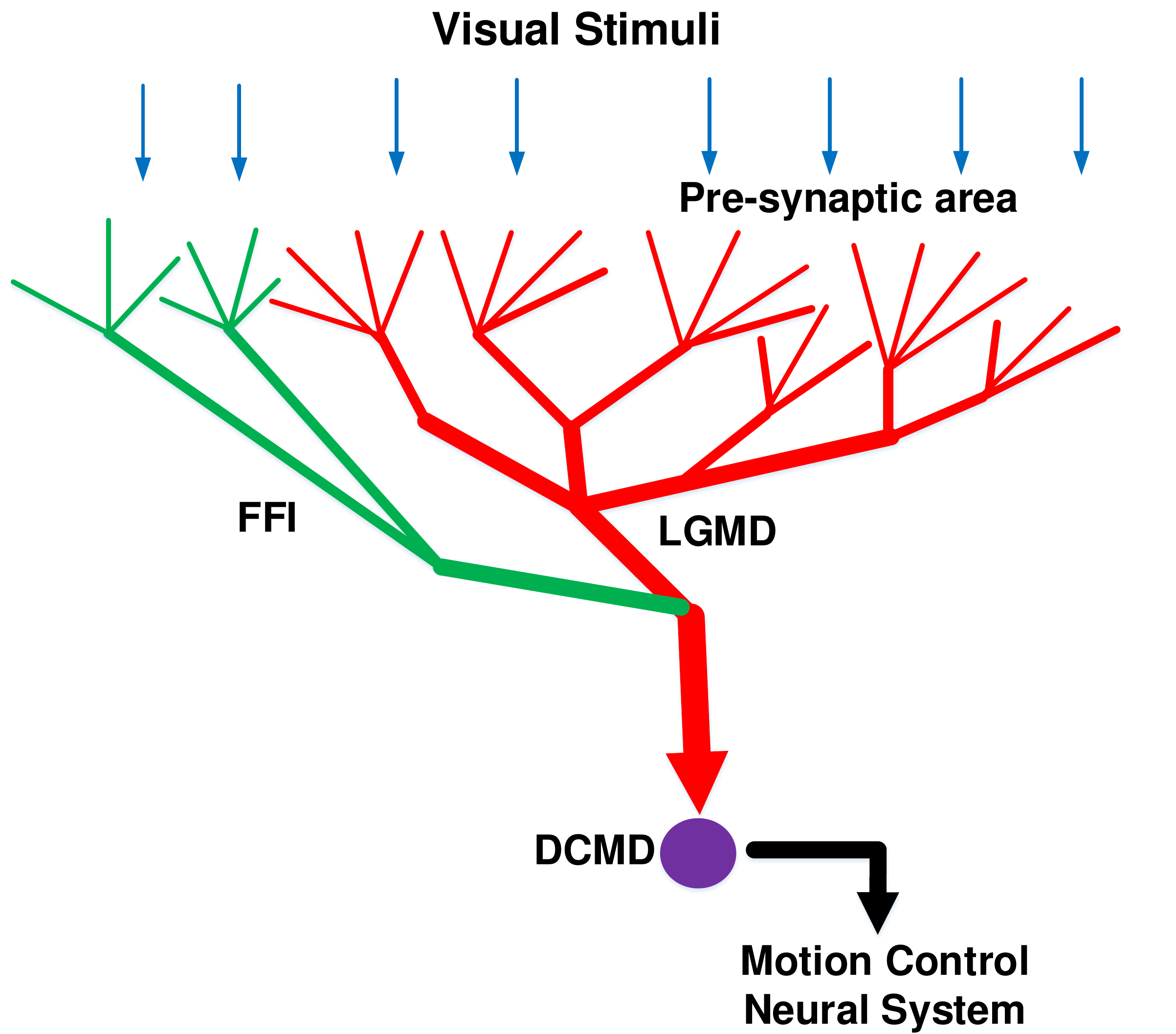}
	\caption{Schematic illustration of the LGMD neuromorphology: the red `dendrites tree' area indicates pre-synaptic visual processing; 
		the green `dendrite trees' field denotes a separate feed-forward inhibitory (FFI) pathway;
		DCMD (descending contra-lateral motion detector) is a one-to-one post-synaptic target neuron conveying the LGMD's spikes to further motion control neural systems.}
	\label{Fig: lgmd-morphology}
	\vspace{-20pt}
\end{figure}

Locusts are well known for fast collision avoidance behaviour on the basis of visual cues. 
A group of lobula giant movement detectors (LGMDs) has been found by biologists (e.g. \cite{LGMDs-2016}). 
The LGMD1 (namely LGMD in this paper) was firstly identified as a moving objects detector and then gradually recognised as a large-field looming sensitive neuron, which responds most vigorously to objects quickly approaching rather than other kinds of movements \cite{LGMDs-2016}. 
The neuromorphology of LGMD is illustrated in Fig. \ref{Fig: lgmd-morphology}.
Such a fascinating neuron has been computationally modelled as collision selective visual neural networks or models (e.g. \cite{LGMD1-Glayer,LGMD1-Silva1,LGMD1-nonlinear,Keil-2004}), and applied in mobile machines like ground robots \cite{Colias-Hu,Fu2017a(LGMDs-IROS),Fu-2018(LGMD1-NN)} and UAVs\cite{UAV2017(LGMD1-spiking)}, and also embodied in hardware implementations like the FPGA \cite{Meng-2010(LGMD1-FPGA)}. 
These works have partially reproduced the LGMD's responses and demonstrated that it features efficient neural computation for quick and reliable looming or collision sensing. 
However, the performance of these methods is still restricted by the background complexity, i.e., the real physical environmental noise including irrelevant motion cues greatly affect the looming detection.

For outdoor vehicle driving scenarios, there are many challenges for artificial collision detection vision systems. 
The more unpredictable and dynamic environments include optical flows caused by ego-motions with the movements of lane markings or rapidly approaching ground shadows, and situations like nearby vehicles approaching or surpassing. 
These are all significantly challenging the performance of visual collision detection systems. 
There have been some modelling studies showing the LGMD's efficacy of collision detection in vehicle driving scenarios. 
For example, a seminal work by Keil et al. demonstrated the effectiveness of ON and OFF mechanisms in an LGMD model for proximity detection in real-world scenes \cite{Keil-2004}. 
Yue et al. introduced a genetic algorithm with an LGMD visual neural network to reduce false collision alerts rate using video sequences from a camera mounted inside a car \cite{LGMD1-car2006}. 
Stafford et al. proposed a method to combine the elementary motion detector (EMD) from the fly visual system with the LGMD for amplifying the colliding and translating stimuli \cite{LGMD1-car2007}. 
More recently, a prominent work was proposed by Harbauer for constructing an LGMD-based collision detection system for vehicles \cite{LGMD-car-2017}. 
In this study, the author specified a `danger zone' in the centre of the vehicle's view to help exclude irrelevant optical flows in the surrounding area.

In this research, on the basis of a recent biological research \cite{LGMDs-2016}, we looked into the signal processing within the pre-synaptic area of the LGMD giant neuron. 
Compared to aforementioned works and our previous model \cite{Fu-2018(LGMD1-NN)}, we propose an adaptive inhibition mechanism, which enables the LGMD to perceive looming cues against complex dynamic background like gratings. 
Importantly, such a mechanism leads to further sharpen up the LGMD's selectivity by rigorously suppressing background shifting and other irrelevant motion including objects receding and translating. 
To verify the effectiveness of this new inhibition mechanism, we investigated its collision detection performance in complex dynamic scenes. 
The rest of this paper is organised as follows:
Section \ref{Section: model} introduces the proposed method; 
Section \ref{Section: experiments} presents our experiments and results, with further discussion; 
this research is summarised in Section \ref{Section: conclusion}.
\section{Model Description}
\label{Section: model}
\begin{figure}[t]
	\centering
	\includegraphics[width=0.8\linewidth]{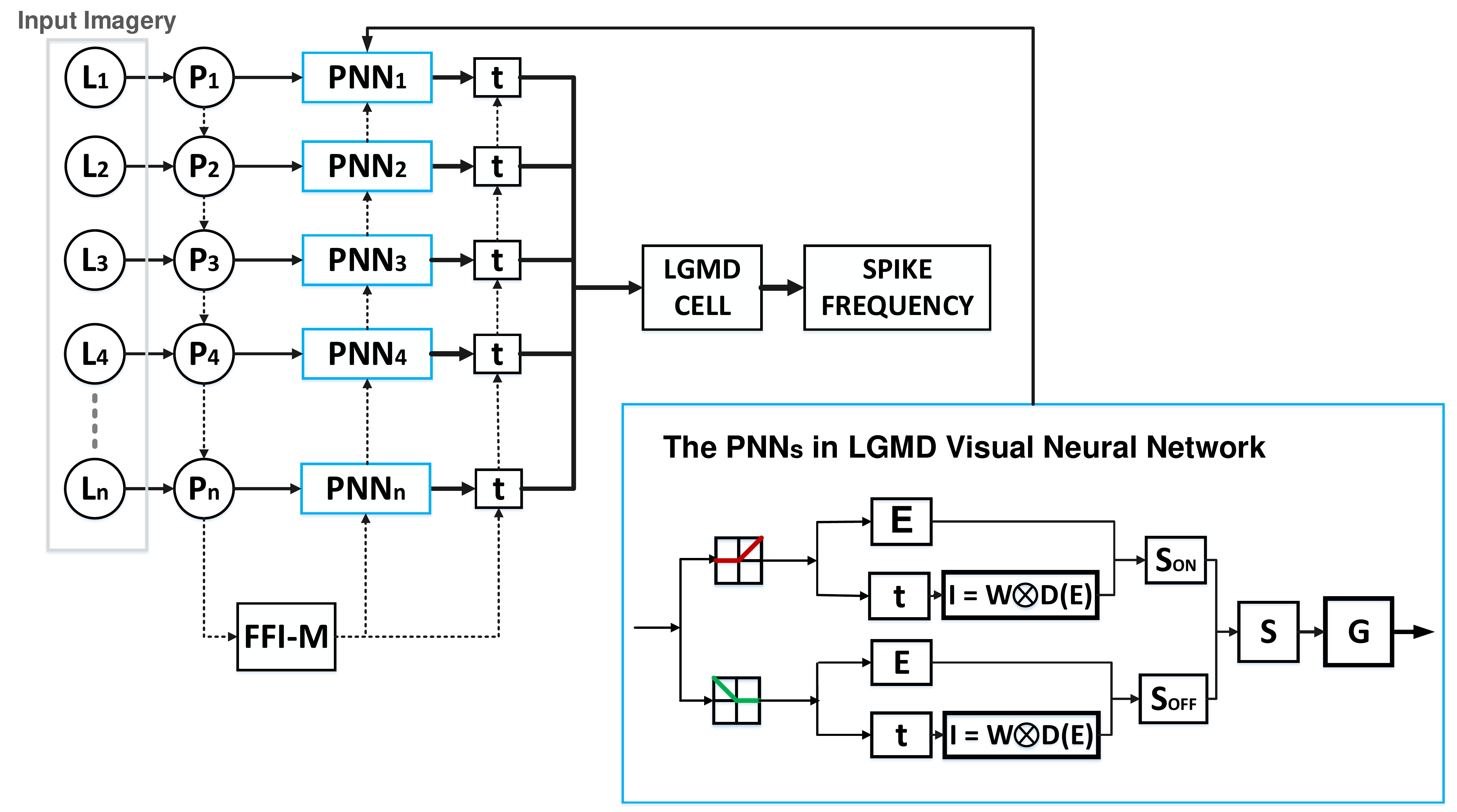}
	\caption{Schematic of the LGMD visual neural network.
		The input is grey-scale imagery.
		Pixel-wise luminance (L) is captured by $n$ number of photoreceptors (P), each relayed into ON and OFF pathways with multi-layers inside partial neural networks (PNN). 
		ON and OFF cells implement the functions of half-wave rectifiers.
		E, I, S and G stand for excitation, inhibition, summation and grouping cells;
		t indicates temporal latency.
		There are spatiotemporal convolutions in the ON/OFF channels. 
		The output is spike frequency.
		A separate FFI-M pathway from the photoreceptor layer adjusts the local inhibition strength and tunes the local excitation latency at every frame.
	}
	\label{Fig: lgmd-neural-network}
	\vspace{-10pt}
\end{figure}

In this section we will present the proposed LGMD visual neural network. 
The neural computation flowchart is illustrated in Fig. \ref{Fig: lgmd-neural-network}.
\subsection{Spatiotemporal Neural Computation in the Pre-synaptic Area}
The first layer of photoreceptors arranged in a 2-D matrix calculates the luminance change between every two successive frames at every local pixel:
\begin{equation}
P(x,y,t) = L(x,y,t) - L(x,y,t-1) + \sum_{i}^{n_p}a_i \cdot P(x,y,t-i).
\label{Eq: p-layer}
\end{equation}
The persistence of luminance change could last for a short while of several ($n_p$) frames. 
The decay coefficient $a_i$ is calculated by $a_i = (1 + e^{u \cdot i})^{-1},$ and $u = 1$.

In addition, from the photoreceptor layer, we also compute the object size dependent feed-forward inhibition (FFI). 
Differently from former LGMD neural networks, e.g.\cite{LGMD1-Glayer,LGMDs-2016,Colias-Hu,Fu-2018(LGMD1-NN)}, where the FFI can directly suppress the LGMD giant neuron if luminance changes rapidly over a large field of view, we propose a new FFI mechanism, namely the feed-forward inhibition mediation (FFI-M), which will be used to tune the local inhibitions and excitations in the ON and OFF pathways. 
We define the output delayed signal as $\hat{F}$, which is computed as:
\begin{equation}
F(t) = \sum_{x}^{R}\sum_{y}^{C}|P(x,y,t)| \cdot n_{cell}^{-1},\quad \frac{d \hat{F}(t)}{d t} = \frac{1}{\tau_{1}}(F(t) - \hat{F}(t)),
\label{Eq: ffi}
\end{equation}
where $C$ and $R$ denote the columns and rows of the photoreceptor layer, and $n_{cell}$ stands for the total number of units, i.e. $n_{cell} = C \times R$. 
The output is delayed by a first-order low-pass filtering with a time constant ($\tau_1 = 10 ms$).

After that, the photoreceptors pass motion information through parallel ON and OFF pathways, depending on luminance increments (ON) or decrements (OFF):
\begin{equation}
P_{on}(x,y,t) = [P(x,y,t)]^{+},\quad P_{off}(x,y,t) = -[P(x,y,t)]^{-},
\label{Eq: half-wave}
\end{equation}
where $[x]^{+}$ and $[x]^{-}$ denote $max(0,x)$ and $min(x,0)$, respectively.

Subsequently, the polarity signals flow into parallel pathways, each possessing multiple layers including excitation (E), inhibition (I) and summation (S).  
Firstly, the ON cell leads the excitation to pass directly to the ON-E layer; meanwhile, it is fed into a low-pass filtering, which gives feedback on delayed information:
\begin{equation}
E_{on}(x,y,t) = P_{on}(x,y,t),\quad \frac{d \hat{E}_{on}(x,y,t)}{d t} = \frac{1}{\tau_{2}}(E_{on}(x,y,t) - \hat{E}_{on}(x,y,t)),
\label{Eq: ON-E}
\end{equation}
where $\hat{E}_{on}$ is the output and $\tau_{2}$ is the latency varying between $60$ and $180ms$. 
The local inhibition in the ON-I layer is then convolved by delayed ON-excitations:
\begin{equation}
I_{on}(x,y,t) = \sum_{i=-1}^{1}\sum_{j=-1}^{1} \hat{E}_{on}(x+i,y+j,t) \cdot W(i+1,j+1),
\label{Eq: ON-I}
\end{equation}
where $W$ denotes a convolution kernel and fits the following matrix:
\begin{equation}
W = \left(
\begin{aligned}
&1/8 &1/4\quad  &1/8\\
&1/4 &1\quad  &1/4\\
&1/8 &1/4\quad  &1/8
\end{aligned}
\right).
\label{Eq: on-kernel}
\end{equation}
It is important to note that the local excitation performs also self-inhibition. 
Next, in the ON-S layer, there is a competition  between the local excitation and inhibition, which represents a linear calculation:
\begin{equation}
S_{on}(x,y,t) = E_{on}(x,y,t) - w \cdot I_{on}(x,y,t),
\label{Eq: localS}
\end{equation}
where $w$ is a local bias to the inhibition. 
Note that only the non-negative values can reach the forthcoming computation. 
In this LGMD neural network, the ON and OFF pathways share the same spatiotemporal neural computation. 
Therefore, for simplicity, we illustrate the processing in the ON pathway only.
After the generation of local ON and OFF excitations in the ON-S and -OFF-S layers, there are interactions between both polarity channels, which represent supralinear computations. 
That is,
\begin{equation}
S(x,y,t) = \theta_1 \cdot S_{on}(x,y,t) + \theta_2 \cdot S_{off}(x,y,t) + \theta_3 \cdot S_{on}(x,y,t) \cdot S_{off}(x,y,t),
\label{Eq: supralinearS}
\end{equation}
where $\{\theta_1$, $\theta_2$, $\theta_3\}$ denote the combination of term coefficients for balancing the contributions of both ON and OFF pathways, all set to $1$ in this model. 
The proposed visual neural network features also a grouping (G) layer for the purpose of reducing noise and for clustering local excitations by expanding edges. 
The G layer processing adopts the methods used in a former LGMD neural network \cite{LGMD1-Glayer}, which is omitted here. 
Moreover, as shown in Fig. \ref{Fig: lgmd-neural-network}, the grouped local excitation has a temporal latency before reaching the LGMD cell. 
The computational role is consistent with a first-order low-pass filtering, with a dynamic time $\tau_g$, updated at every frame (initially set to $10 ms$).
\subsection{Adaptive Inhibition Mechanism}
In the proposed LGMD visual neural network, the FFI-M pathway is crucial to adjust the local biases to both the ON-inhibition and the OFF-inhibition in the PNN, at every frame. 
That is,
\begin{equation}
w = max \left( \sigma_{1}, \frac{\hat{F}(t)}{T_{f}} \right),\quad \hat{\tau_{g}} = \tau_{g} \cdot max \left( \sigma_{2}, 1 - \frac{\hat{F}(t)}{T_{f}} \right),
\label{Eq: ffi-m}
\end{equation}
where $T_{f} = 20$ is a threshold and $\sigma_{1} = 0.5$. 
This mechanism works effectively to make the neural network adapt to different levels of background complexity. 
More precisely, the model is inhibited by dramatic changes of background clutter, like background shifting and grating movements. 
Most importantly, the model can still detect looming objects within dynamic background clutter.  
In addition, the FFI-M pathway tunes the temporal latency of grouped local excitations, where the time delay at every frame is updated by a coefficient compared to a very small real number $\sigma_{2}=0.01$. 
This dynamic temporal tuning indicates that the time latency of local excitations reaching the LGMD will become shorter as the objects growing on the field of view, i.e., the looming case.
\subsection{The LGMD Cell}
Following the pre-synaptic visual processing, the LGMD giant neuron pools all the local excitations forming the neural potential, which is then exponentially transformed to the sigmoid membrane potential as follows:
\begin{equation}
k(t) = \sum_{x}^{R}\sum_{y}^{C}G(x,y,t),\quad K(t) = \left(1 + e^{-k(t) \cdot (n_{cell} \cdot \sigma_{3})^{-1}}\right)^{-1},
\label{Eq: smp}
\end{equation}
where the coefficient $\sigma_{3} = 1$ shapes the neural potential within $[0.5,1)$. 
$G$ is the grouped local excitation in the G layer (Fig. \ref{Fig: lgmd-neural-network}). 
After that, we apply a spike frequency adaptation mechanism, which contributes to further sharpen up the looming selectivity via weakening the LGMD's responses to translating and receding stimuli. 
The corresponding computation can be found in \cite{Fu-2018(LGMD1-NN)}.
\subsection{Output Spike Frequency}
The membrane potential is then exponentially mapped to spikes by an integer-valued function:
\begin{equation}
S^{spike}(t) = \left[e^{(\sigma_{4} \cdot (K(t) - T_{sp}))}\right],
\label{Eq: spiking}
\end{equation}
where $T_{sp}=0.7$ indicates the spiking threshold and $\sigma_{4}=10$ is a scale parameter affecting the firing rate at every frame. 
After that, we compute the spike frequency (spikes per second) within a range of discrete frames as the model output to indicate the recognition of collisions or not:
\begin{equation}
Coll(t) = \left\{
\begin{aligned}
&\text{True},\ \text{if}\ \left(\sum_{i=t-N_{ts}}^{t}S^{spike}(i)\right) \cdot 1000 / (N_{ts} \cdot \tau_{i}) \ge T_{sf}\\
&\text{False},\ \text{otherwise}
\end{aligned}
\right.,
\label{Eq: collision}
\end{equation}
where $N_{ts}=6$ denotes the amount of frames constituting a short time window and $T_{sf}=15 \sim 30\ spikes/s$ is the spike frequency threshold. 
$\tau_{i}$ stands for the discrete time interval in milliseconds between successive frames.
\section{Experimental Evaluation}
\label{Section: experiments}
%
All the experiments can be categorised into two types of tests:
we firstly tested the proposed LGMD visual neural network against synthetic movements in dynamic background clutter, with various spatiotemporal sinusoidal grating movements and shifting of a panoramic natural scene; 
we then investigated its performance in very complex vehicle driving scenarios consisting of many crash and near-miss scenes.

The simulations input to the proposed model were greys-scale videos with $30 Hz$ sampling frequency. 
The vehicle video sequences were adapted from dashboard camera recordings at $18 Hz$ \cite{youtube-crashes}. 
The resolutions are $380 \times 334$, $540 \times 270$, $352 \times 288$ for grating, panoramic and vehicle videos, respectively.
\subsection{Synthetic Visual Stimuli Testing}
\begin{figure}[t]
	\centering
	\subfloat{\includegraphics[width=0.4\linewidth]{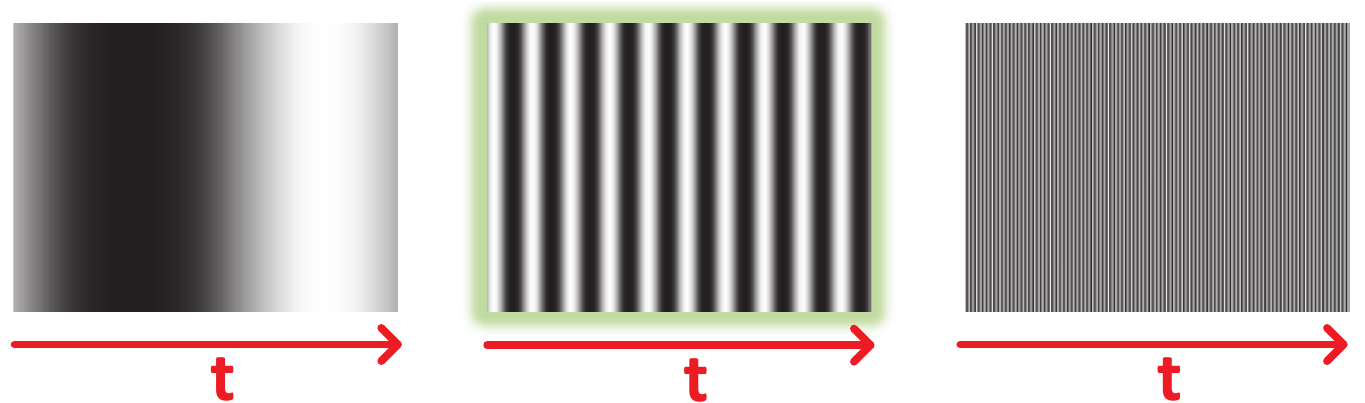}}
	\vfil
	\vspace{-0.15in}
	\subfloat{\includegraphics[width=0.6\linewidth]{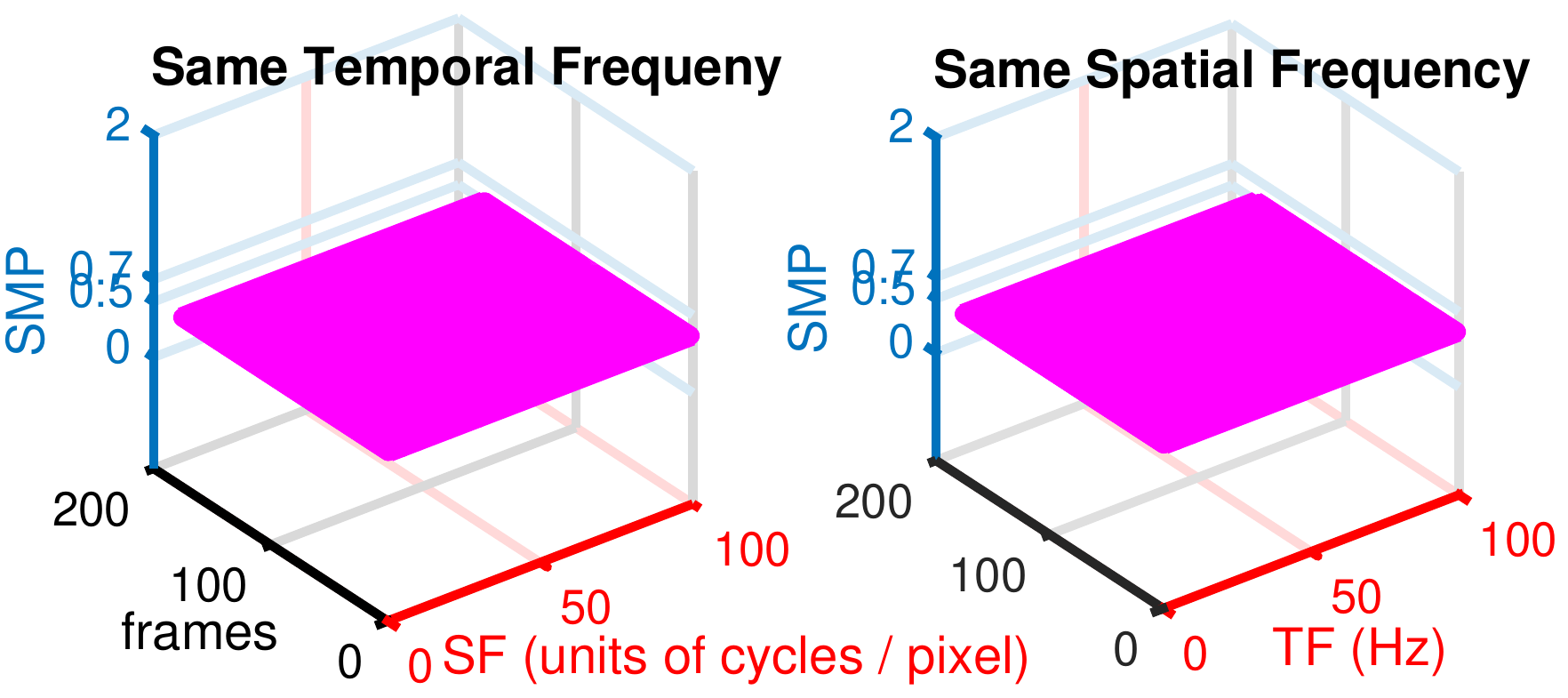}}
	\caption{Membrane potentials of the proposed LGMD model by sinusoidal gratings with a wide range of spatial (SF) and temporal (TF) frequencies:
		the firing threshold is set at \textbf{0.7}; The potentials at $0.5$ denote non-response of the LGMD model.}
	\label{Fig: exp: simu-grating}
	\vspace{-10pt}
\end{figure}

In the first type of experiments, we aim to demonstrate that the proposed LGMD model can detect objects approaching in complex dynamic backgrounds. 
Firstly, as basic trials for biologically visual systems, we tested it against grating movements with a wide range of spatial and temporal frequencies. 
As shown in Fig. \ref{Fig: exp: simu-grating}, challenged by grating movements alone, the proposed LGMD is completely inhibited during each course. 
The results verify that the proposed adaptive inhibition mechanism works effectively to deal with grating movements, which well reconciles with the responses of a biological LGMD.
\begin{figure}[t]
	\vspace{-20pt}
	\centering
	\subfloat{\includegraphics[width=0.45\linewidth]{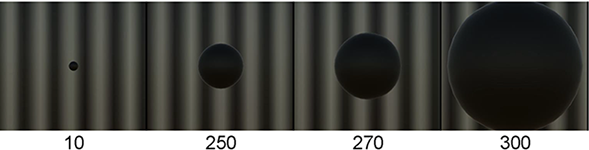}}
	\hfil
	\hspace{0.2in}
	\subfloat{\includegraphics[width=0.45\linewidth]{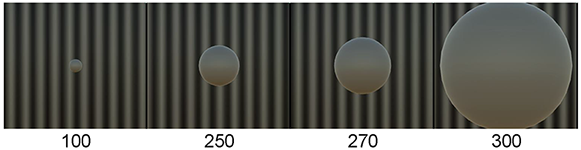}}
	\vfil
	\vspace{-0.1in}
	\subfloat{\includegraphics[width=0.49\linewidth]{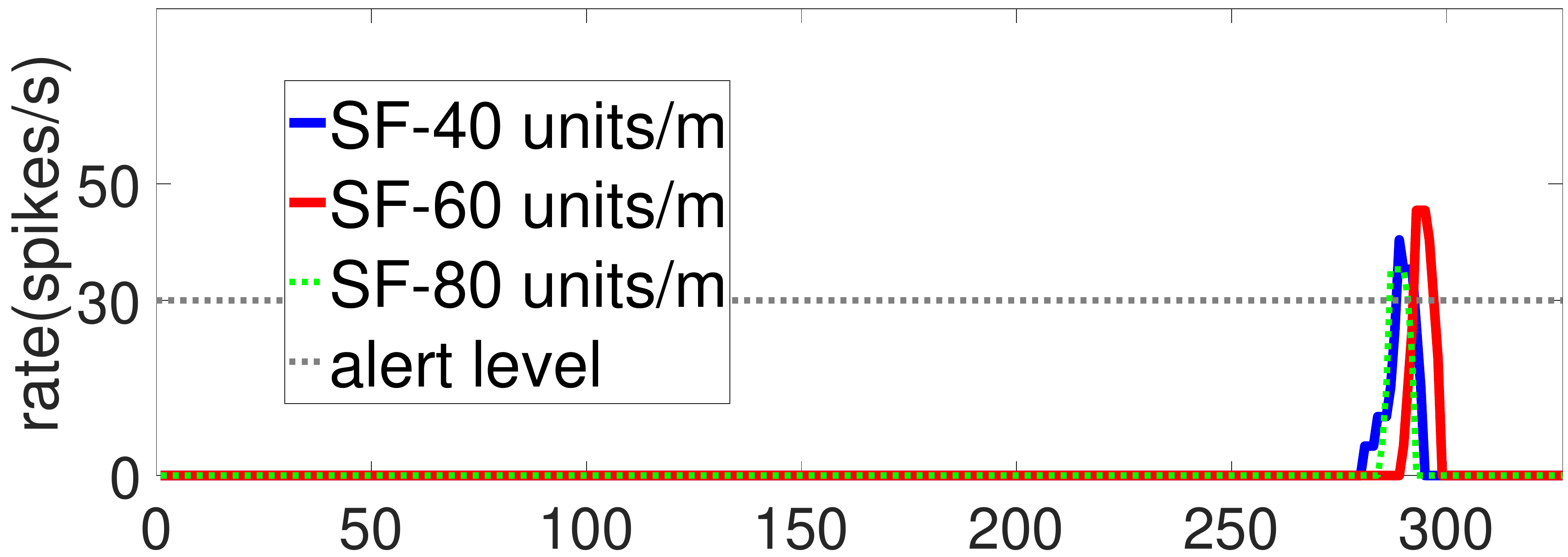}}
	\hfil
	\subfloat{\includegraphics[width=0.49\linewidth]{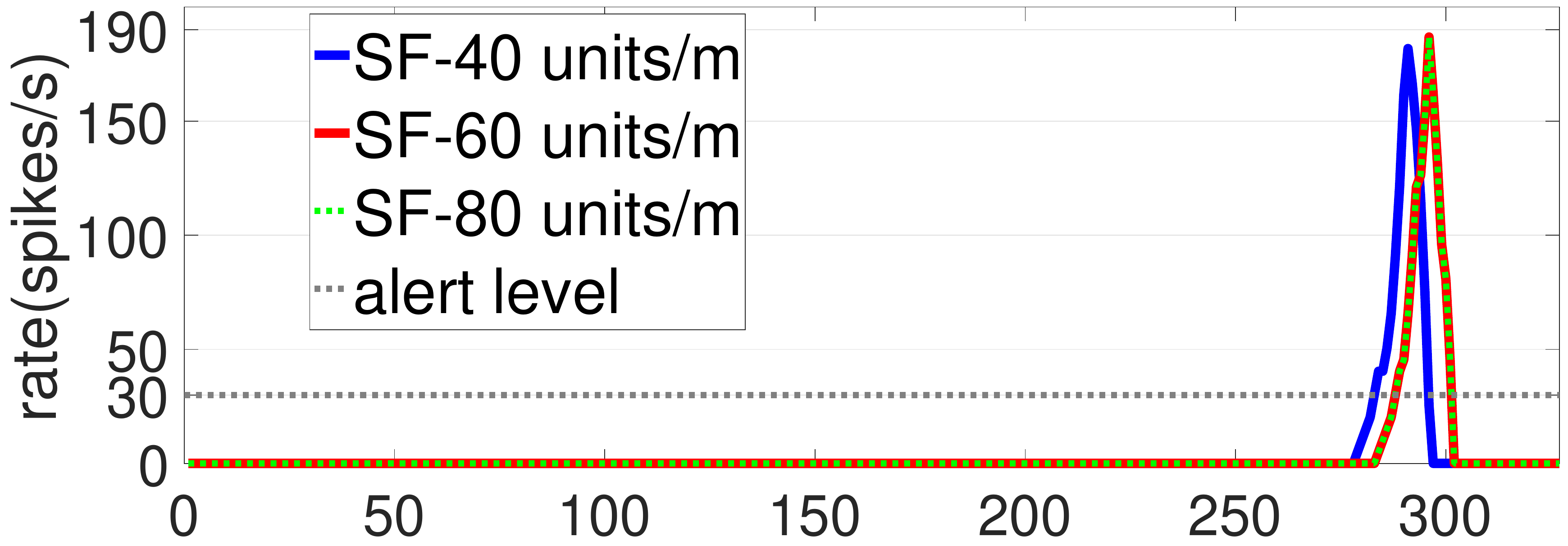}}
	\caption{Spike frequency of the proposed LGMD by dark and light objects looming in grating stimuli with three different SF. 
		The alert level is set at \textbf{30} spikes/s. 
	}
	\label{Fig: exp: looming-in-grating}
	\vspace{-10pt}
\end{figure}
\begin{figure}[!h]
	\centering
	\subfloat{\includegraphics[width=0.45\linewidth]{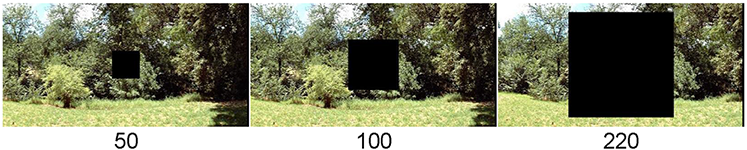}}
	\hfil
	\hspace{0.2in}
	\subfloat{\includegraphics[width=0.45\linewidth]{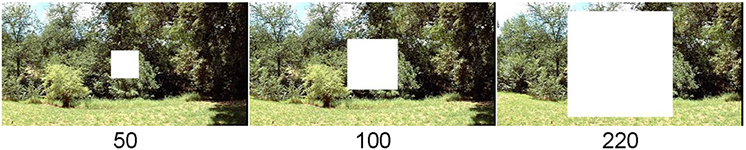}}
	\vfil
	\vspace{-0.1in}
	\subfloat{\includegraphics[width=0.49\linewidth]{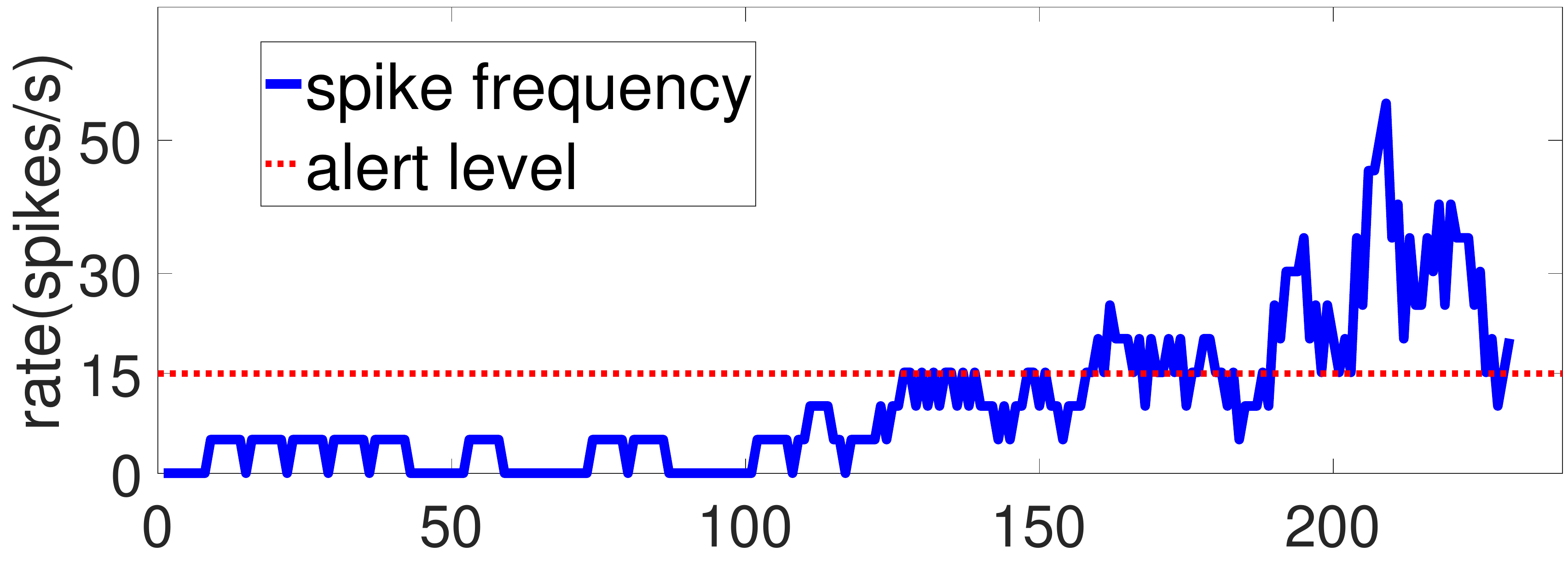}}
	\hfil
	\subfloat{\includegraphics[width=0.49\linewidth]{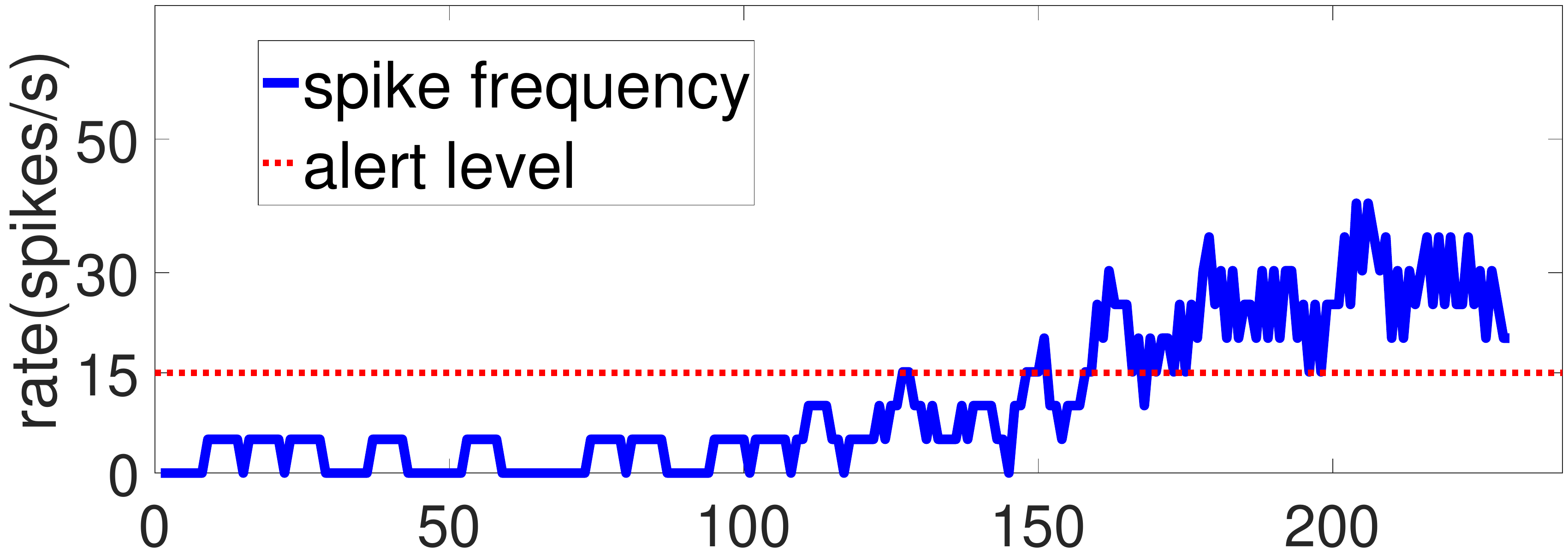}}
	\vfil
	\vspace{-0.1in}
	\subfloat{\includegraphics[width=0.45\linewidth]{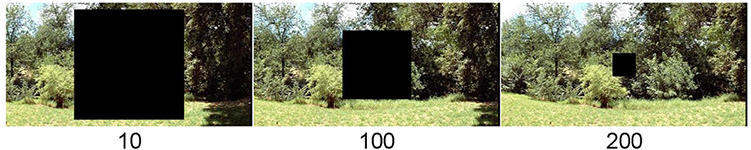}}
	\hfil
	\hspace{0.2in}
	\subfloat{\includegraphics[width=0.45\linewidth]{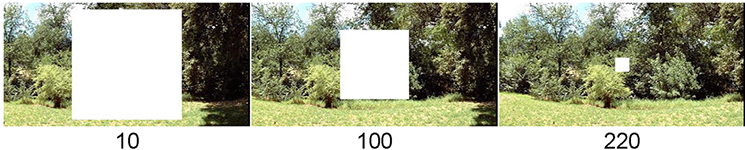}}
	\vfil
	\vspace{-0.1in}
	\subfloat{\includegraphics[width=0.49\linewidth]{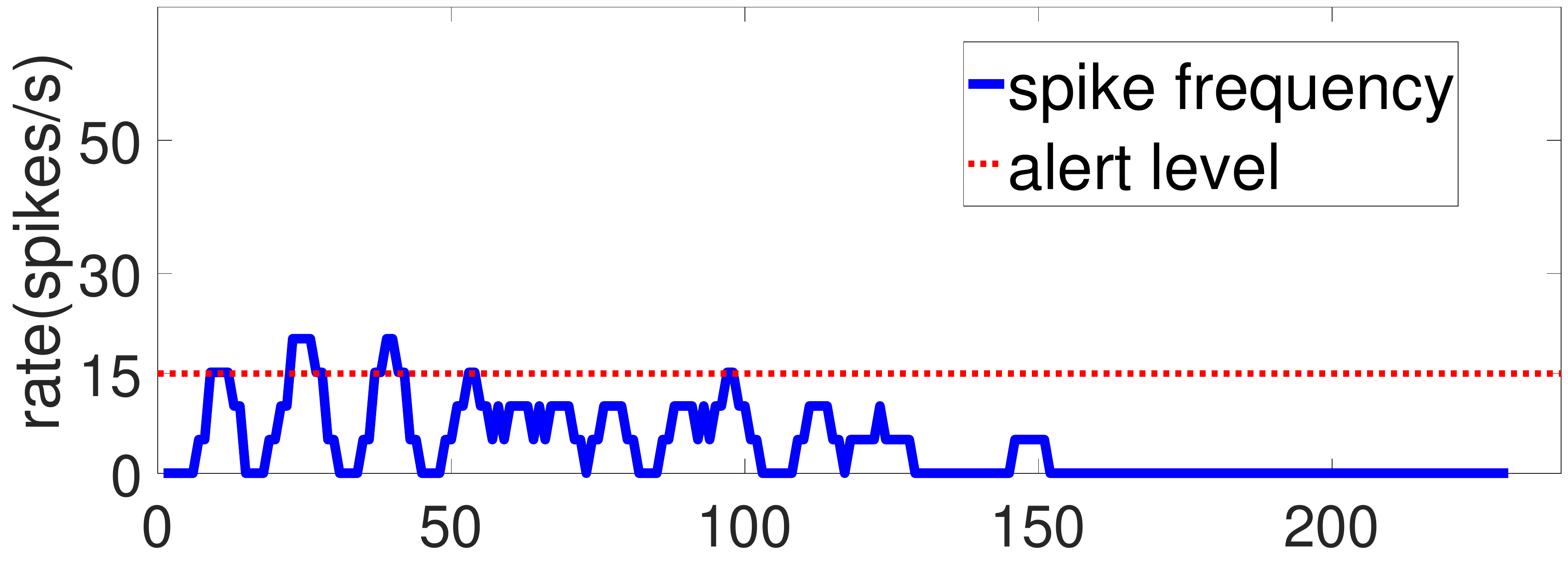}}
	\hfil
	\subfloat{\includegraphics[width=0.49\linewidth]{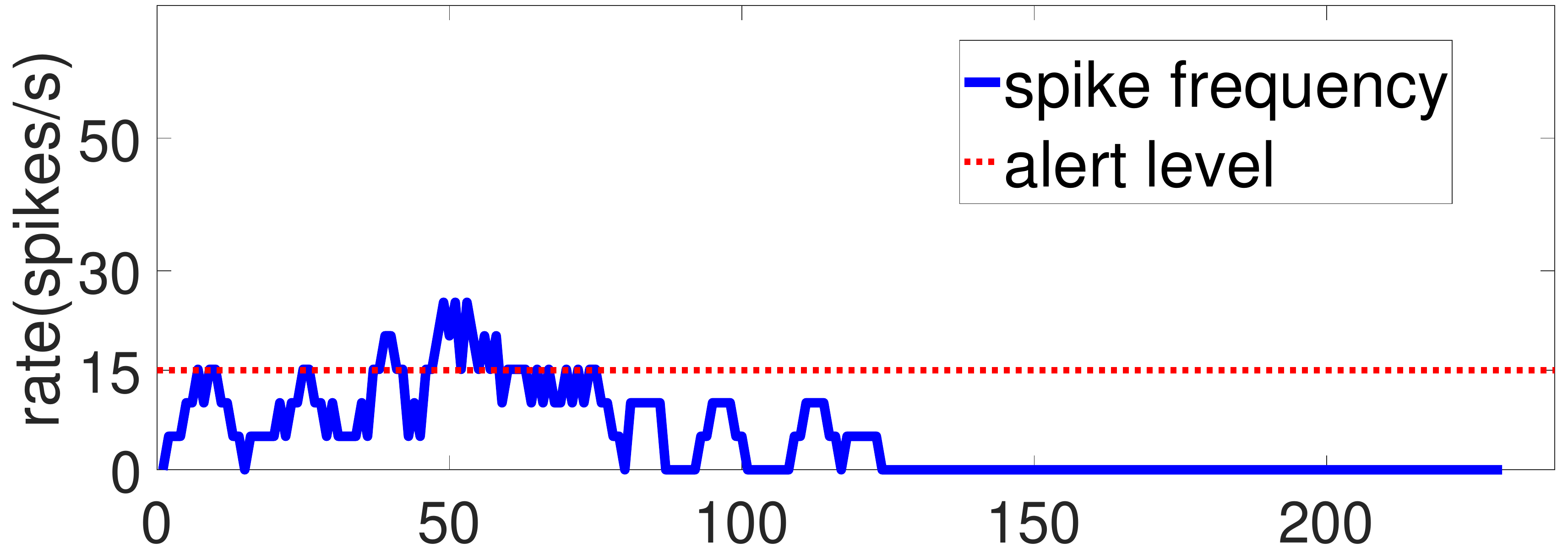}}
	\vfil
	\vspace{-0.1in}
	\subfloat{\includegraphics[width=0.45\linewidth]{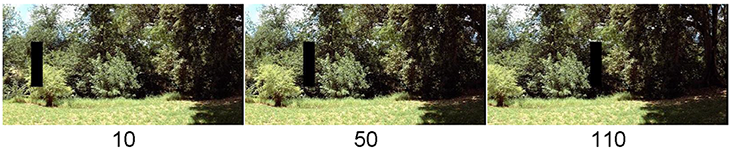}}
	\hfil
	\hspace{0.2in}
	\subfloat{\includegraphics[width=0.45\linewidth]{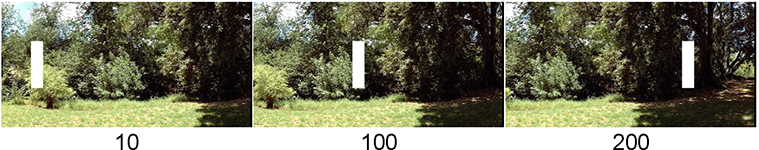}}
	\vfil
	\vspace{-0.1in}
	\subfloat{\includegraphics[width=0.49\linewidth]{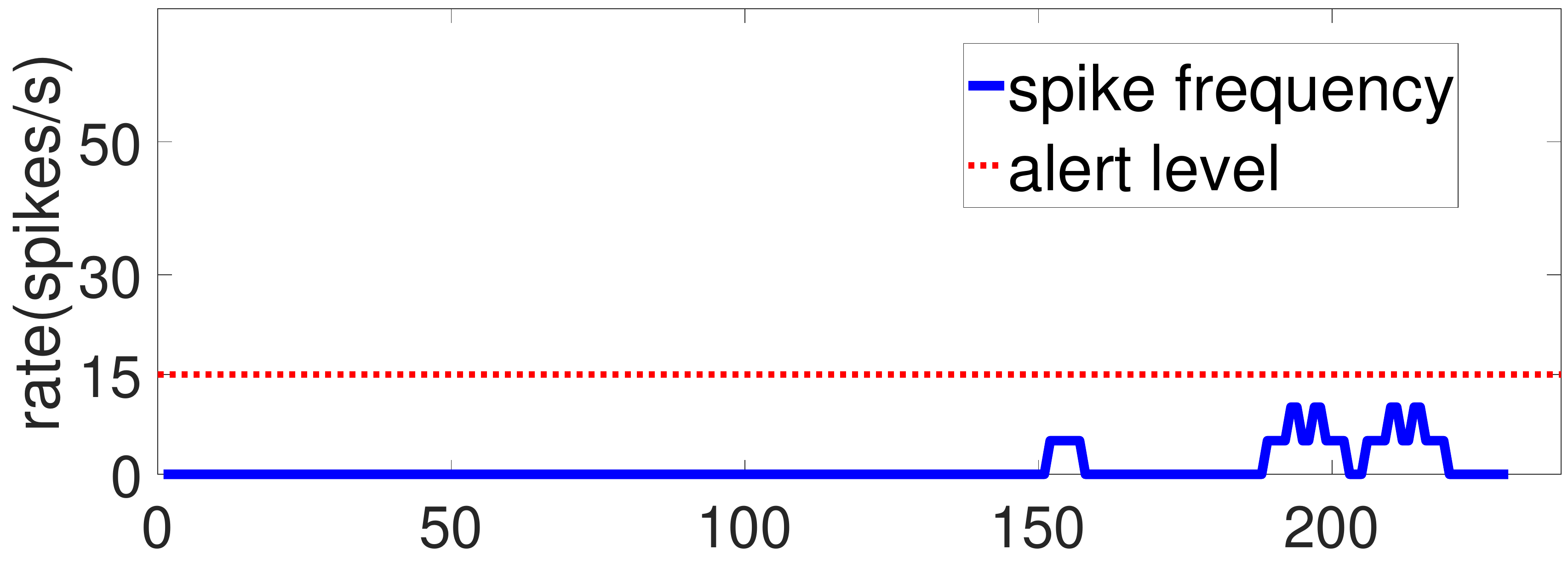}}
	\hfil
	\subfloat{\includegraphics[width=0.49\linewidth]{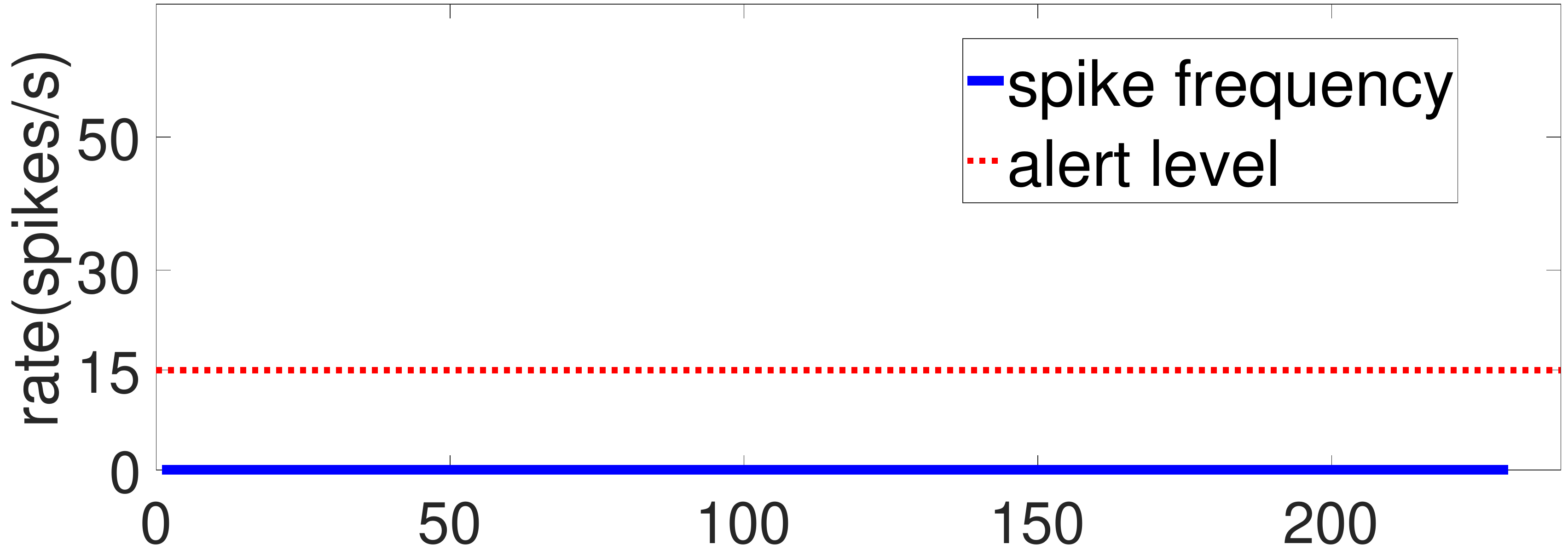}}
	\caption{Spike frequency of the proposed LGMD by dark and light objects approaching, receding and translating embedded in shifting natural background. 
		The alert level is set at \textbf{15} spikes/s.
	}
	\label{Fig: exp: movements-in-nature}
	\vspace{-20pt}
\end{figure}

Subsequently, we simulated dark and light objects looming against grating movements in the background. 
In this case, the object approaches or moves away from the field of view at a constant linear speed of $10.8\ cm/s$; the spatial frequency varies at $40$, $60$, $80\ units/m$. 
The results in Fig. \ref{Fig: exp: looming-in-grating} demonstrate that the proposed LGMD can robustly recognise either dark or light object approaching within grating stimuli regardless of the background grating frequencies. 
The model represents dramatically increasing spike frequencies near the end of approaching when the object gets close to the retina. 
Notably, the light looming object generates a much higher spiking rate, which indicates the LGMD's sensitivity to the contrast between looming objects and background, i.e, the LGMD responds more strongly to looming objects with larger contrast.
\begin{figure}[H]
	\centering
	\subfloat{\includegraphics[width=0.48\linewidth]{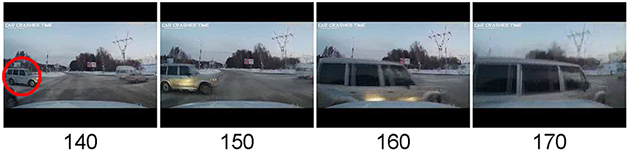}}
	\hfil
	\subfloat{\includegraphics[width=0.48\linewidth]{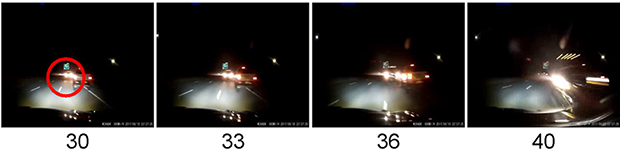}}
	\vfil
	\vspace{-0.15in}
	\subfloat{\includegraphics[width=0.49\linewidth]{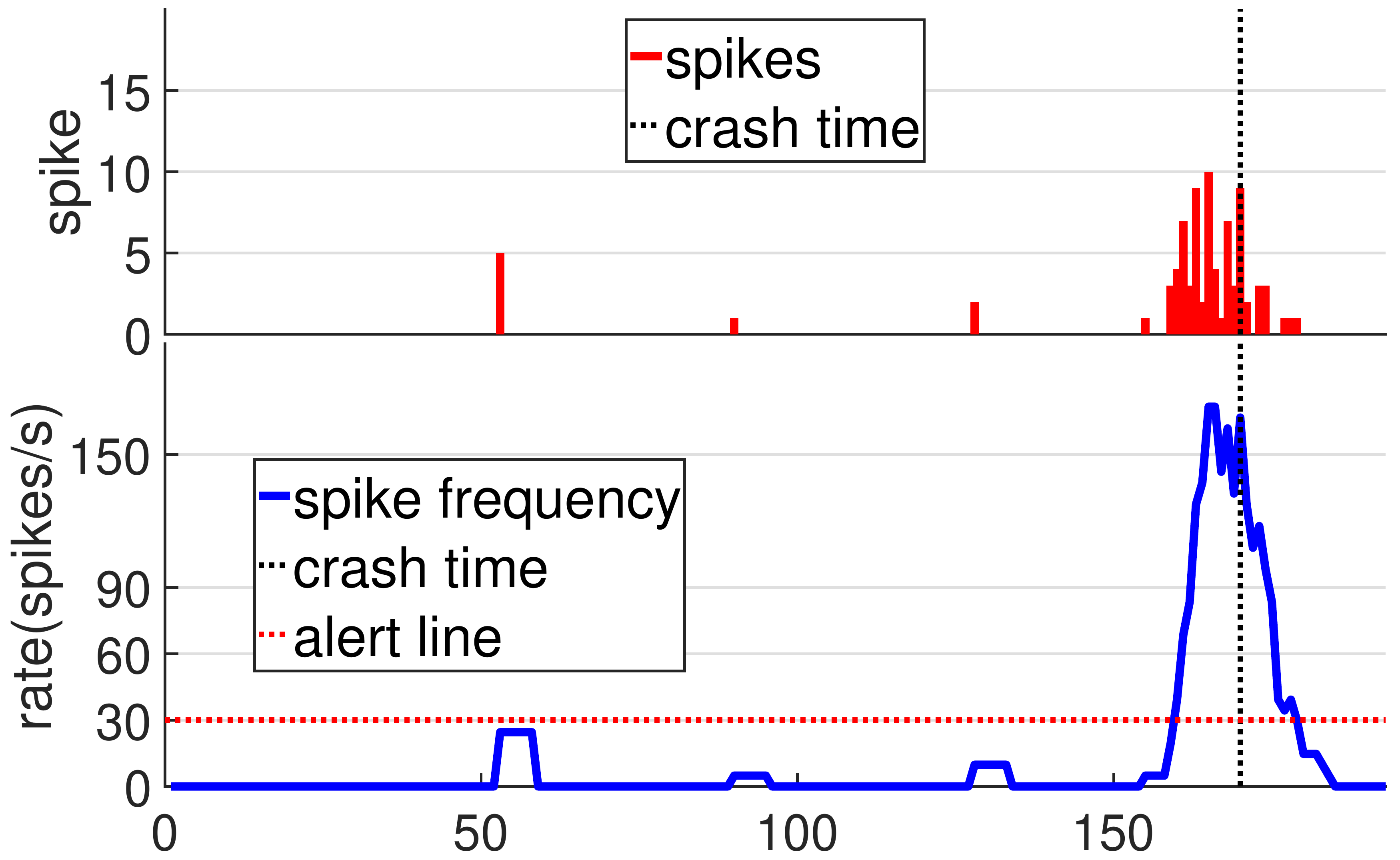}}
	\hfil
	\subfloat{\includegraphics[width=0.49\linewidth]{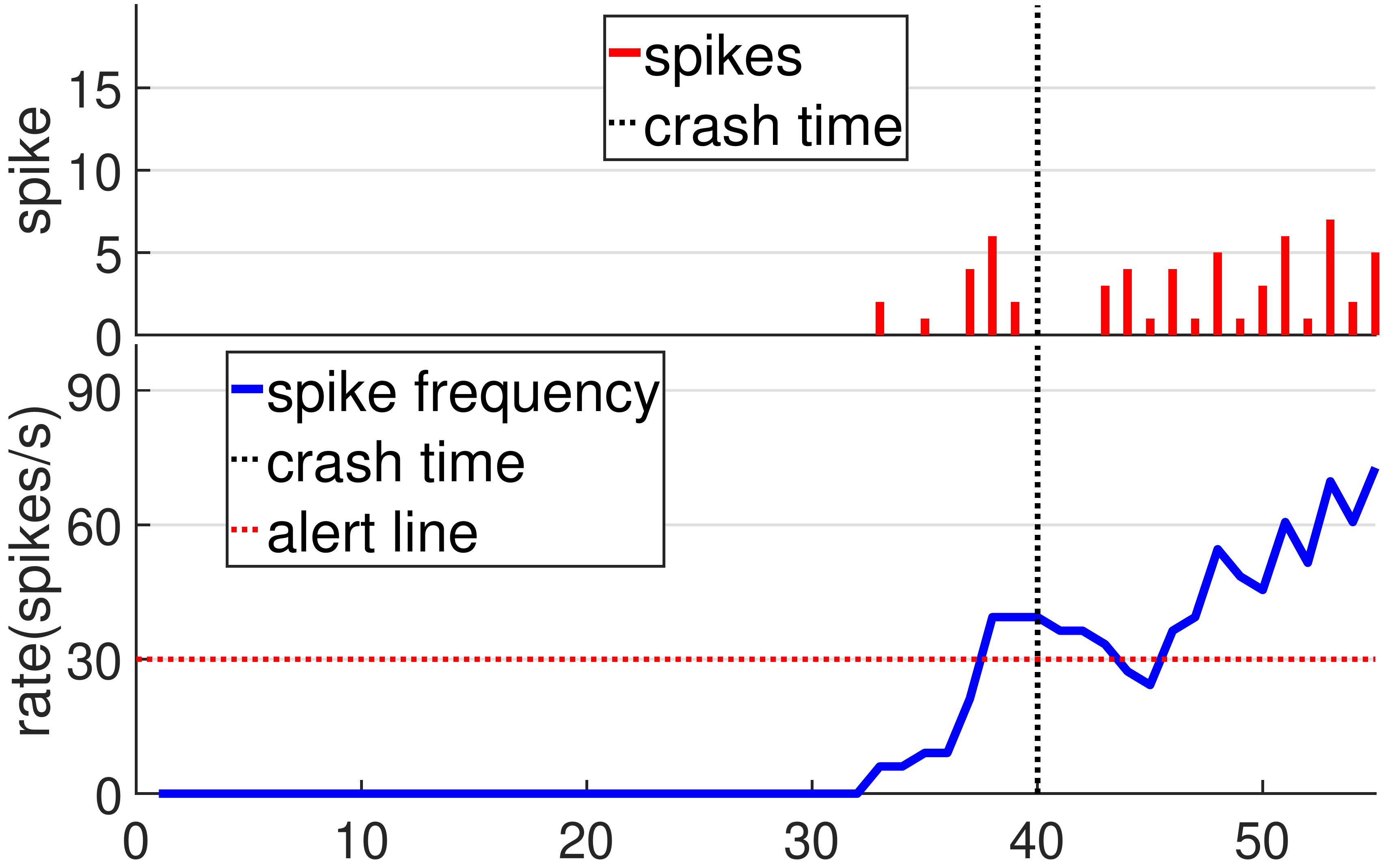}}
	\vfil
	\vspace{-0.1in}
	\subfloat{\includegraphics[width=0.48\linewidth]{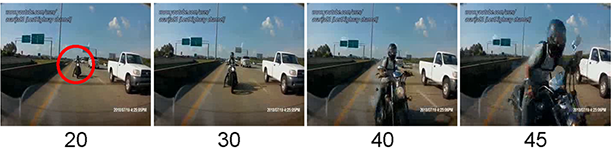}}
	\hfil
	\subfloat{\includegraphics[width=0.48\linewidth]{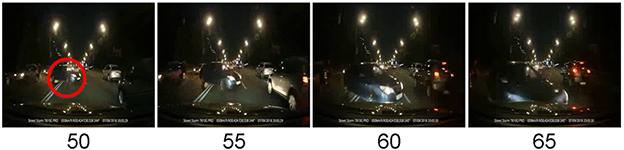}}
	\vfil
	\vspace{-0.15in}
	\subfloat{\includegraphics[width=0.49\linewidth]{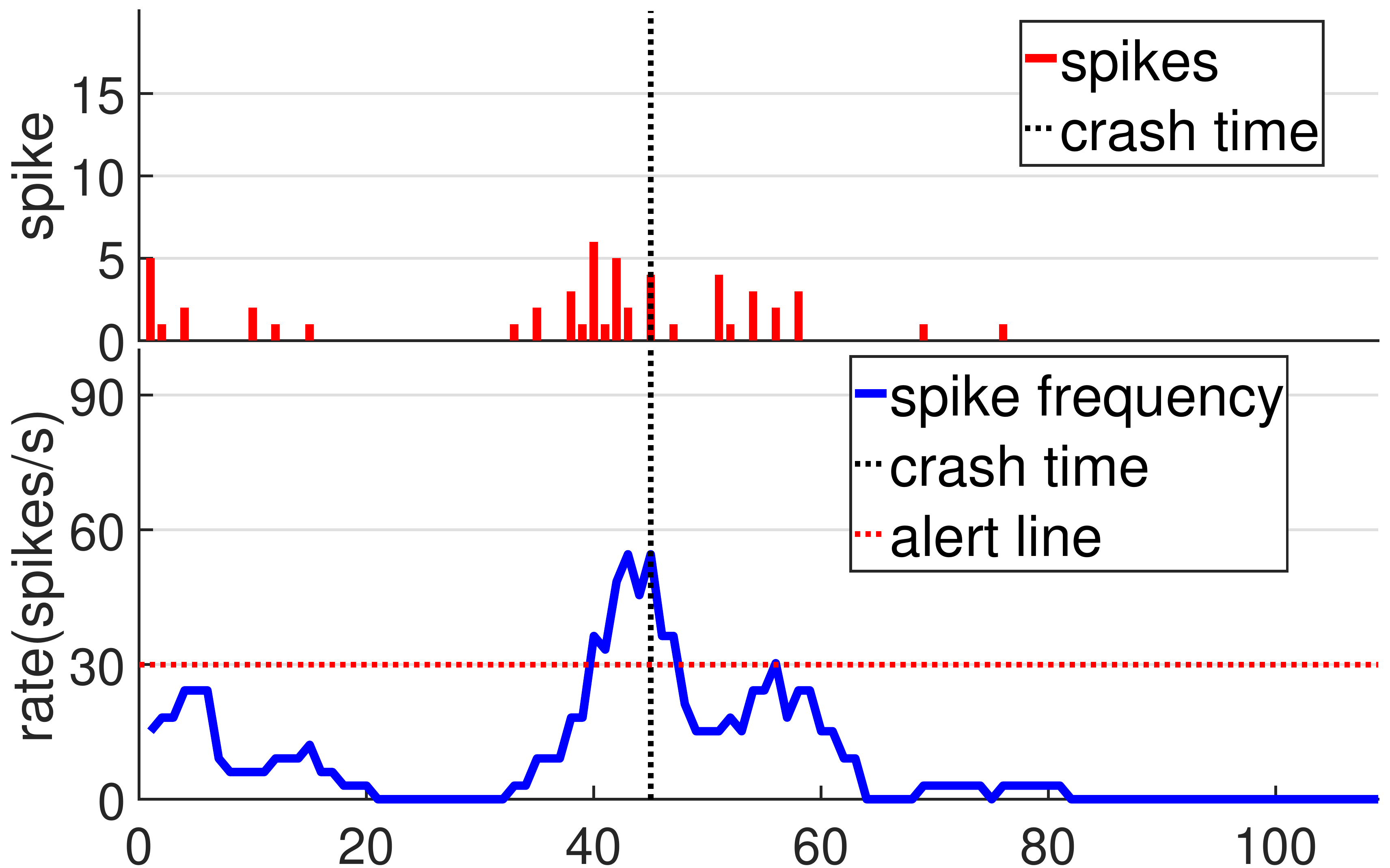}}
	\hfil
	\subfloat{\includegraphics[width=0.49\linewidth]{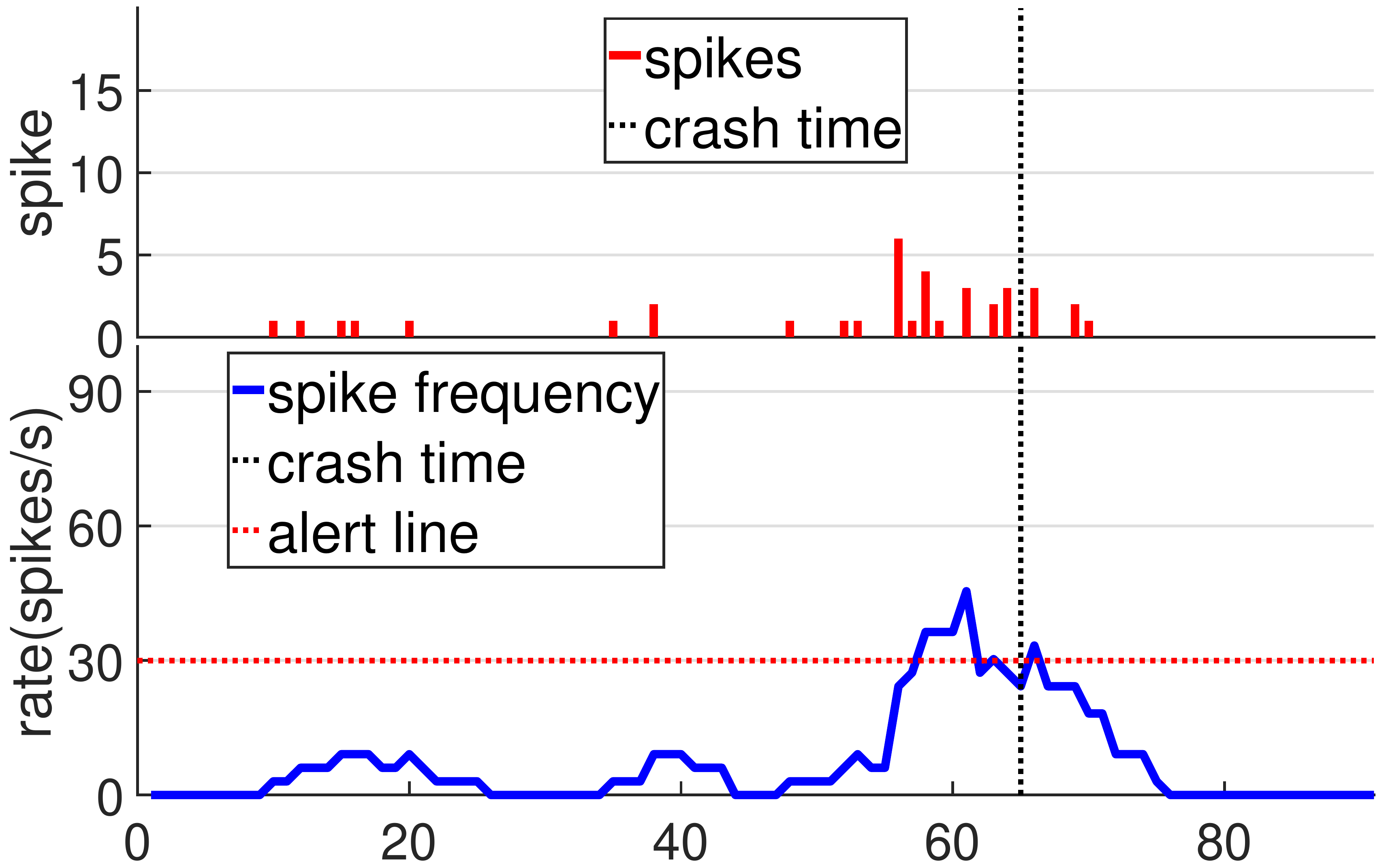}}
	\vfil
	\vspace{-0.1in}
	\subfloat{\includegraphics[width=0.48\linewidth]{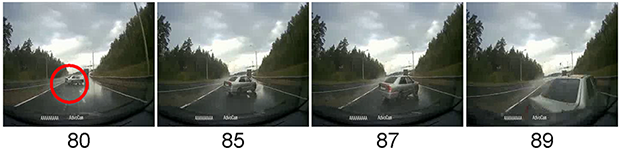}}
	\hfil
	\subfloat{\includegraphics[width=0.48\linewidth]{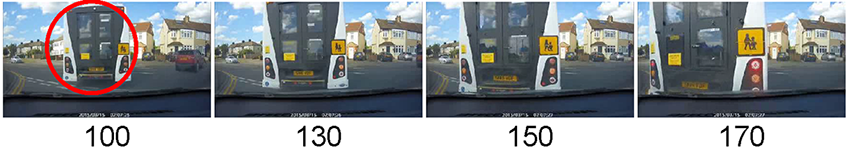}}
	\vfil
	\vspace{-0.15in}
	\subfloat{\includegraphics[width=0.49\linewidth]{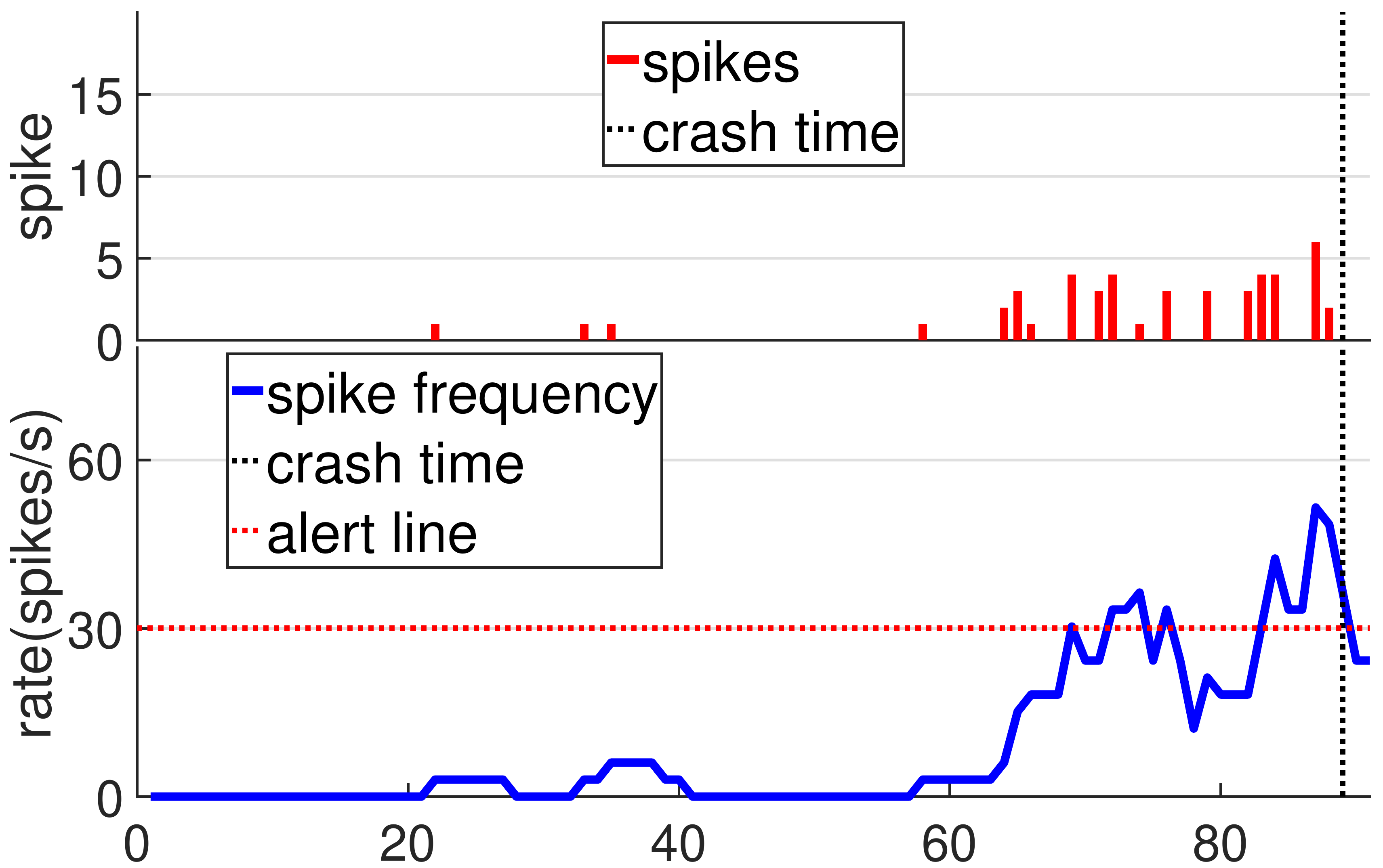}}
	\hfil
	\subfloat{\includegraphics[width=0.49\linewidth]{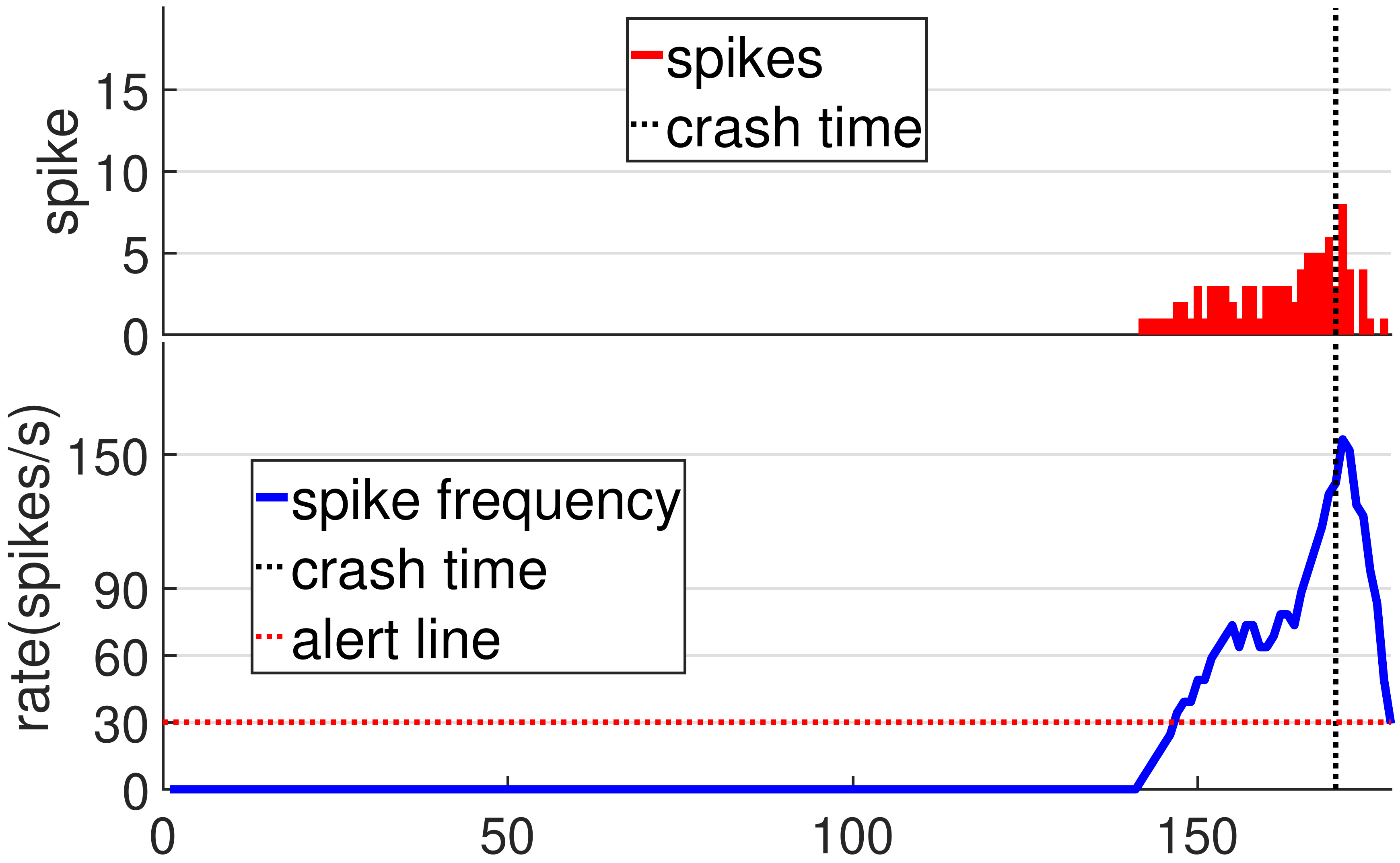}}
	\vfil
	\vspace{-0.1in}
	\subfloat{\includegraphics[width=0.48\linewidth]{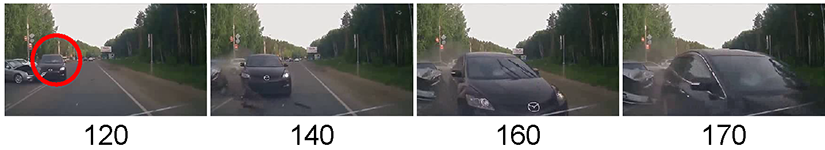}}
	\hfil
	\subfloat{\includegraphics[width=0.48\linewidth]{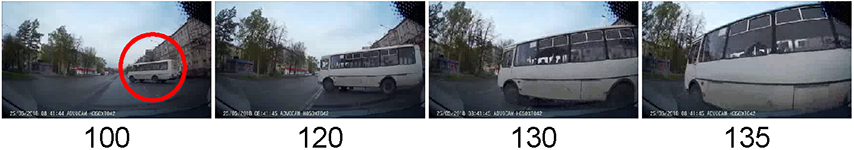}}
	\vfil
	\vspace{-0.15in}
	\subfloat{\includegraphics[width=0.49\linewidth]{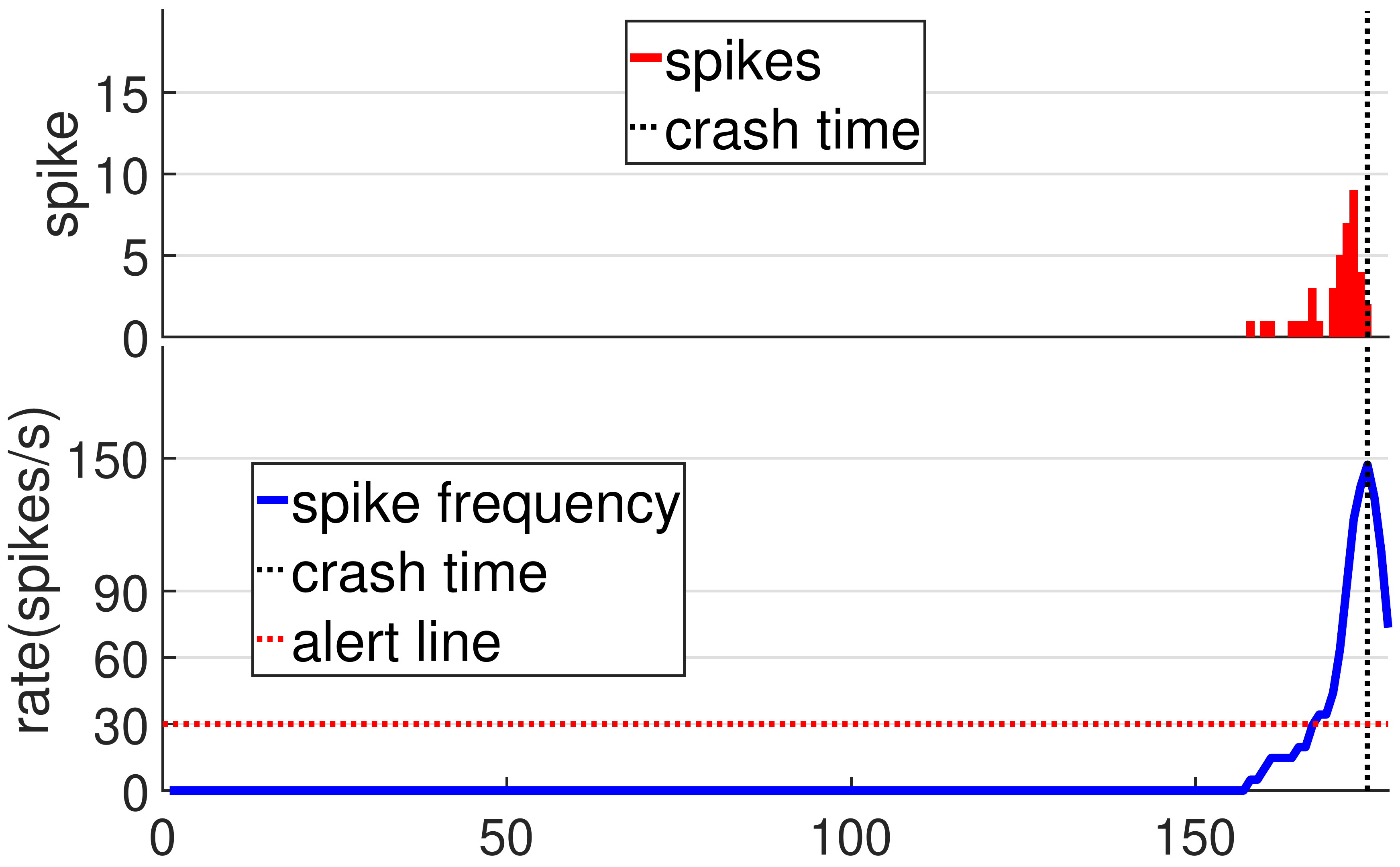}}
	\hfil
	\subfloat{\includegraphics[width=0.49\linewidth]{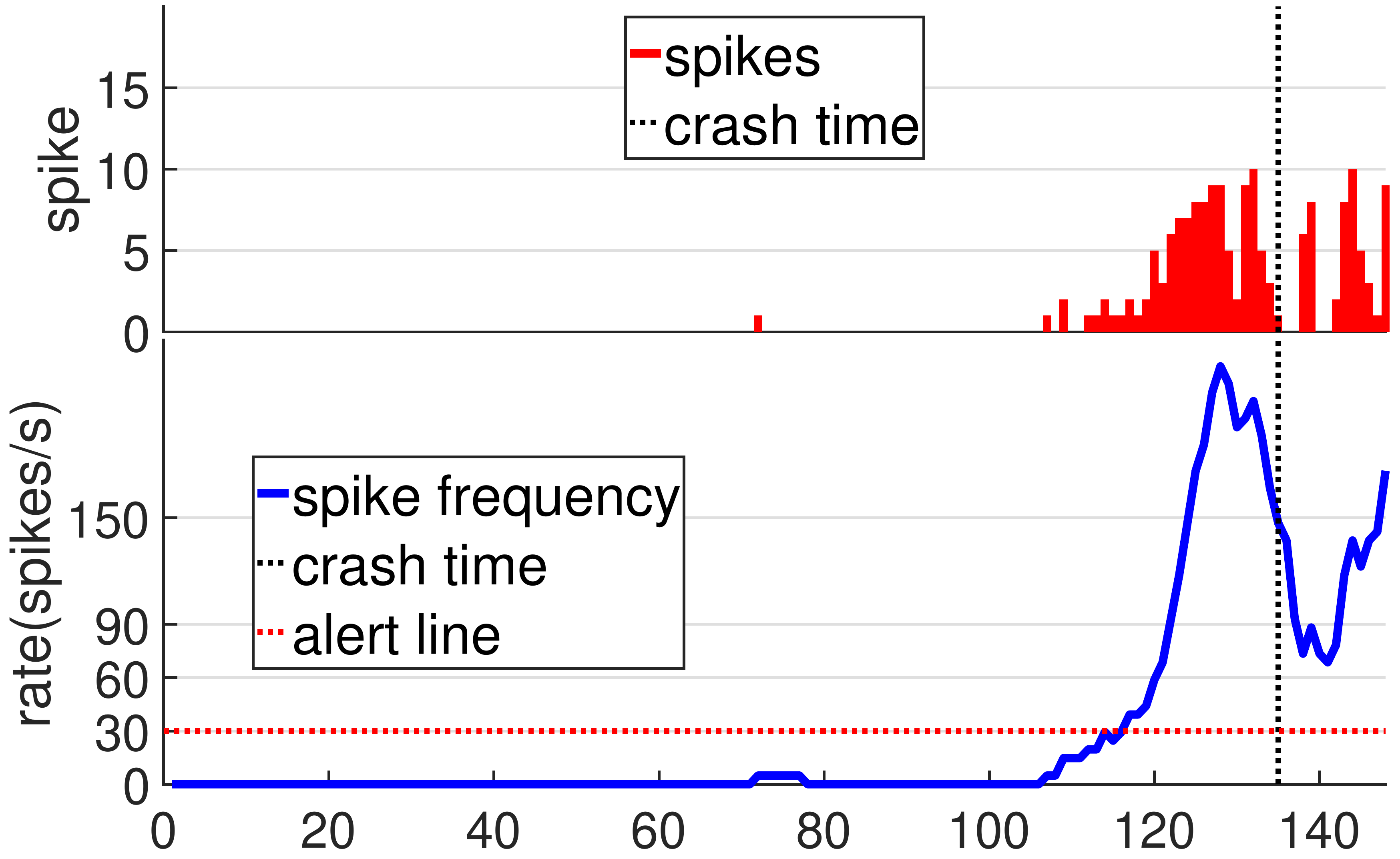}}
	\caption{Spike frequency of the proposed LGMD by vehicle crash scenarios. 
		The spikes and the frequencies are both depicted. 
		The vertical dashed lines indicate the real `crash moments'. 
		A red circle marks the potential colliding object in the snapshots.
	}
	\label{Fig: exp: vehicle-crashes}
\end{figure}

In the last part of the synthetic stimuli experiments, we simulated movements including looming, recession and translation embedded in the shifting of a panoramic natural scene. 
As illustrated in Fig. \ref{Fig: exp: movements-in-nature}, the proposed LGMD responds most strongly to dark and light looming stimuli compared to receding and translating ones. 
The spike frequency gradually increases when looming objects grow on the field of view, and then exceeds the collision alert level remaining high for a long period. 
On the other hand, the LGMD shows lower spike frequency at the start of the object recession only, and little or no response to translations. 
The results verify the effectiveness of the proposed LGMD model for extracting only looming cues from a dynamic cluttered background.
\vspace{-10pt}
\subsection{Real-world Driving Scenes Testing}
In the second type of experiments, the proposed neural network was tested in much more complex vehicle driving scenes, which comprise mixed movements to interfere the detection of real `dangers'. 
Our goal was to investigate its performance for fast collision perception on vehicles. 
We categorised the vehicle driving video sequences adapted from \cite{youtube-crashes} into two types of scenarios, i.e., the fatal crashes and the near-miss scenes.

In the first case, as depicted in Fig. \ref{Fig: exp: vehicle-crashes}, the proposed neural system works effectively in perceiving imminent collisions very quickly, either in daylight or night driving scenarios. 
More specifically, the spike frequency increases very quickly and goes beyond the alert threshold before the highlighted real `crash moments'. 
Otherwise, the LGMD maintains a low-rate rate below the threshold. 
Perceiving dangers before crashes is of great importance for vehicles with or without drivers. 
The proposed LGMD is highly activated before crashes, so it can be a reliable assistant alert system to improve collision avoidance during navigation.

\begin{figure}[t]
	\vspace{-10pt}
	\centering
	\subfloat{\includegraphics[width=0.48\linewidth]{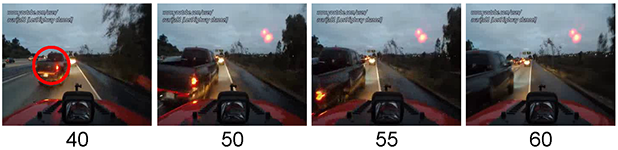}}
	\hfil
	\subfloat{\includegraphics[width=0.48\linewidth]{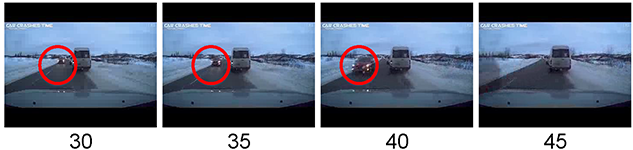}}
	\vfil
	\vspace{-0.15in}
	\subfloat{\includegraphics[width=0.49\linewidth]{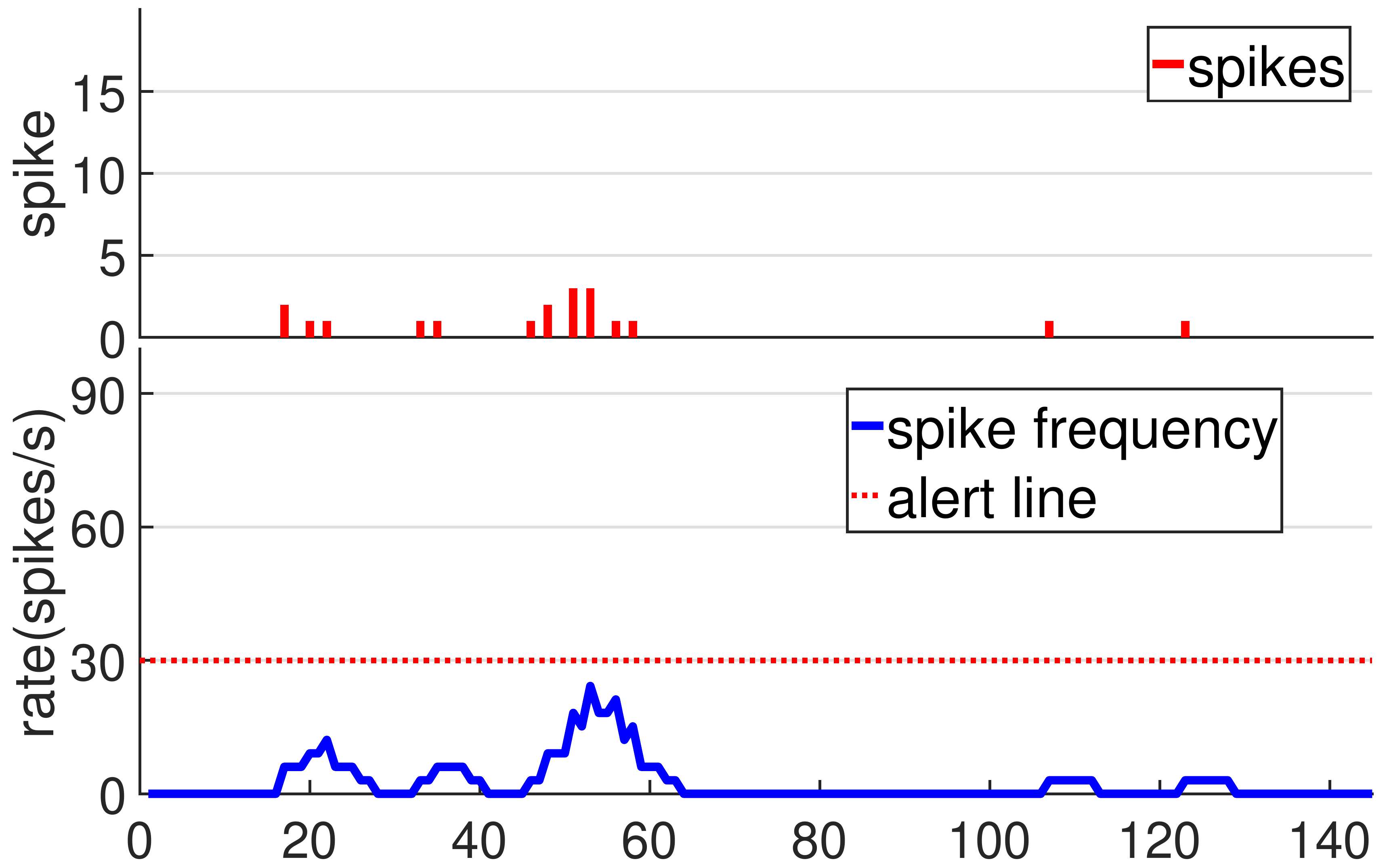}}
	\hfil
	\subfloat{\includegraphics[width=0.49\linewidth]{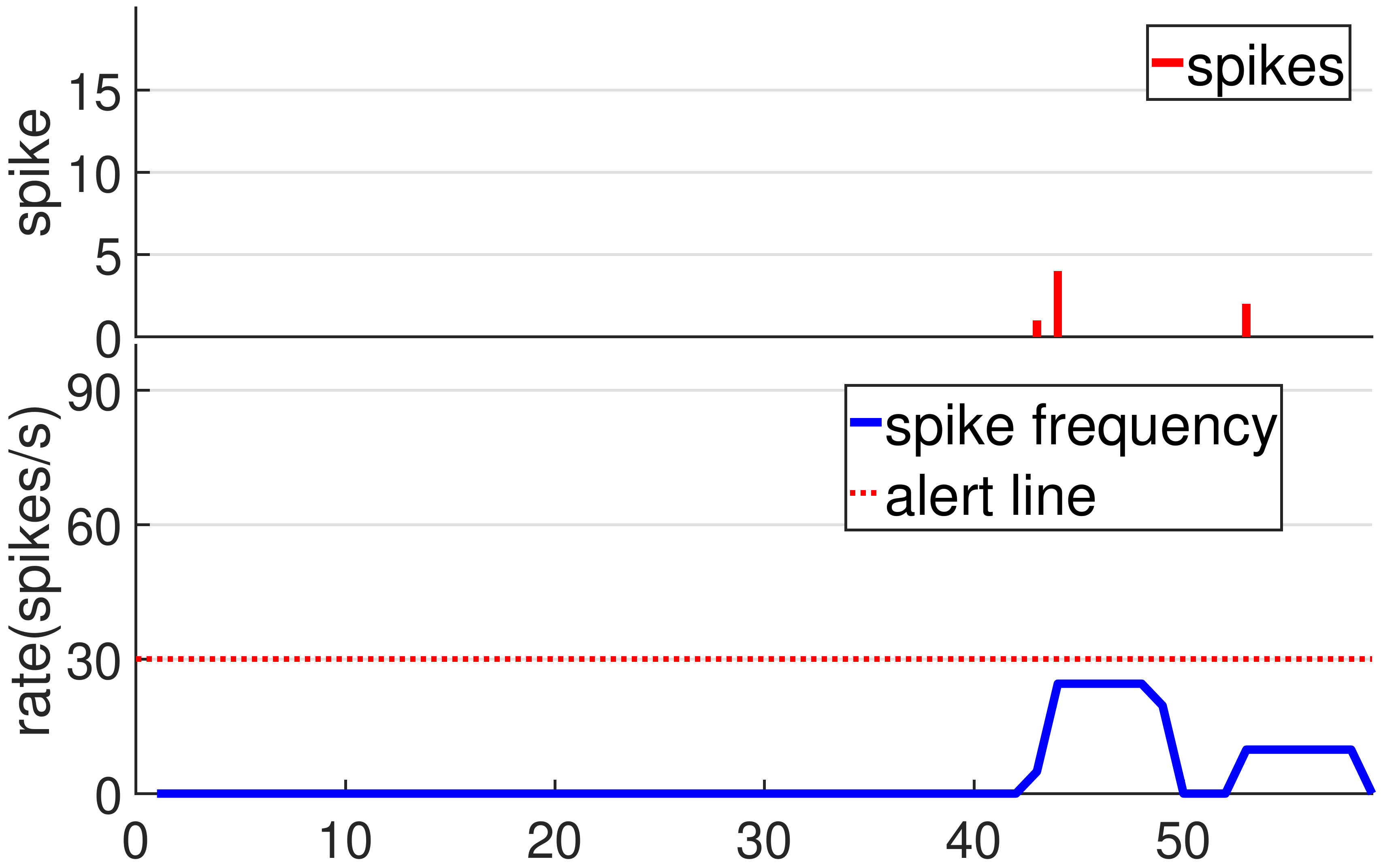}}
	\vfil
	\vspace{-0.1in}
	\subfloat{\includegraphics[width=0.48\linewidth]{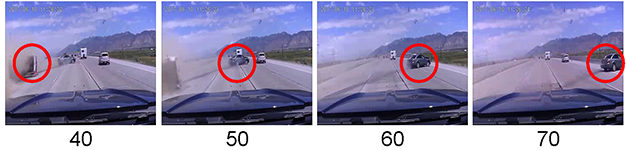}}
	\hfil
	\subfloat{\includegraphics[width=0.48\linewidth]{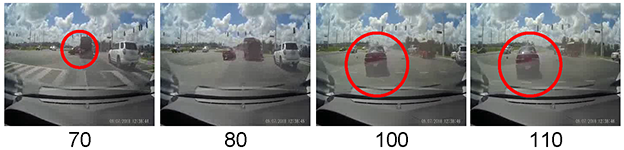}}
	\vfil
	\vspace{-0.15in}
	\subfloat{\includegraphics[width=0.49\linewidth]{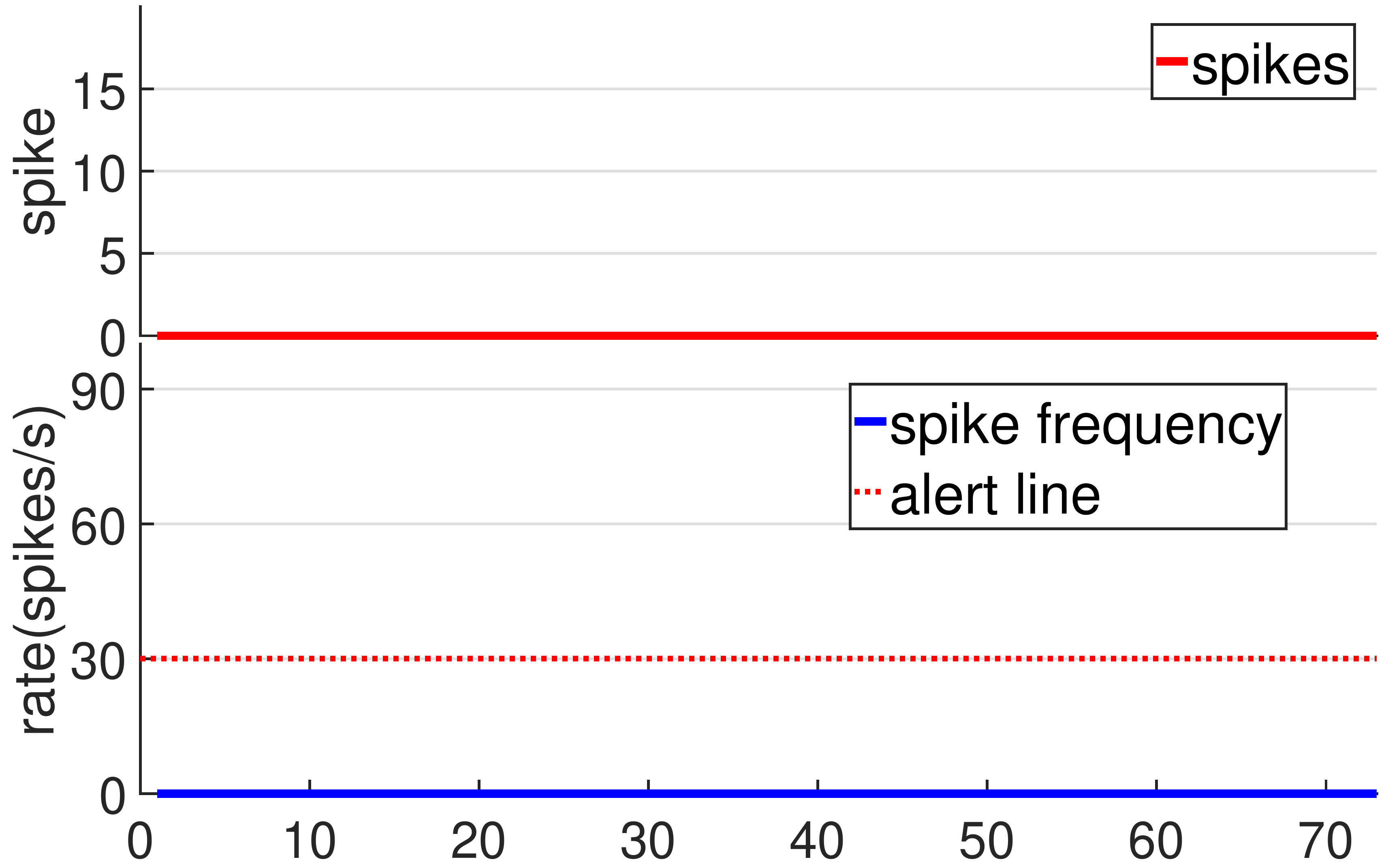}}
	\hfil
	\subfloat{\includegraphics[width=0.49\linewidth]{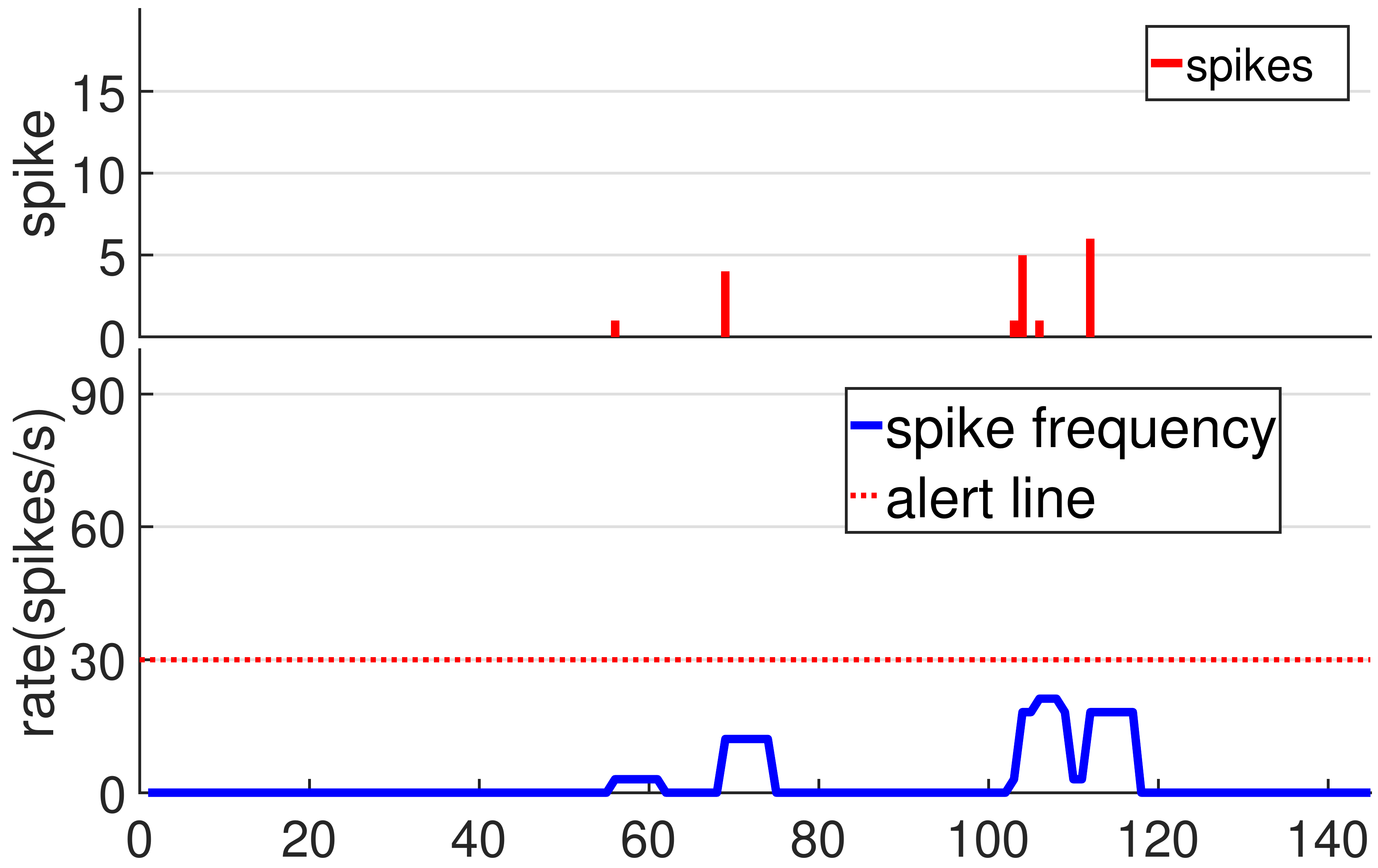}}
	\caption{Spike frequency of the proposed LGMD by vehicle near-miss scenes.
	}
	\label{Fig: exp: vehicle-nearmiss}
	\vspace{-10pt}
\end{figure}

For comparison, we also challenged the proposed neural system with some near-miss situations where the vehicle avoided the collision, or faced nearby-lane approaching and translating vehicles. 
As illustrated in Fig. \ref{Fig: exp: vehicle-nearmiss}, compared to the crash cases, the proposed model generates few sparse spikes or remains quiet. 
The results demonstrate the proposed LGMD visual neural network can well discriminate urgent collisions by rapid approaching objects from other irrelevant end less dangerous situations.
\vspace{-10pt}
\subsection{Discussion}
Building upon our previous work, we have shown the efficacy of an adaptive inhibition mechanism in the LGMD visual neural network dealing with looming perception in complex dynamic scenes. 
Notably, we showed its robust collision perception ability in vehicle driving scenarios including fatal crashes. 
However, there are still challenges to be solved by the proposed method. 
Indeed, although it can perceive impending crashes from the frontal view, we found that nearby surpassing or approaching objects may affect the collision detection performance by causing false alerts. 
In this case, a single neuron computation is insufficient, whereas the integration of multiple neural systems could be more effective. 
In addition, the alert firing threshold, which is now manually defined, should be also adaptive to varying complexity of environmental dynamics.
\section{Concluding Remarks}
\label{Section: conclusion}
This paper has introduced a bio-plausible visual neural network inspired by the locust looming sensitive giant neuron -- the LGMD -- for fast collision perception in complex dynamic scenes, including vehicle driving scenarios. 
Compared to previous studies, we focused on an adaptive inhibition mechanism capable of dealing with different levels of background complexity. 
The experimental results verified the feasibility and robustness of the proposed method in potential real-world applications. 
To improve road safety, the proposed model could be a good collision detection system embodied in miniaturised sensors for future autonomous vehicles and robots.
%
%
%
%
%
\bibliographystyle{splncs04}
\bibliography{qinbing-aiai}
\end{document}